\documentclass[11pt]{article}

\usepackage[final]{acl}

\usepackage{times}
\usepackage{latexsym}

\usepackage[T1]{fontenc}

\usepackage[utf8]{inputenc}

\usepackage{microtype}

\usepackage{inconsolata}

\usepackage{graphicx}
\usepackage{tabularx}
\usepackage{booktabs}
\usepackage{amsmath}
\usepackage{listings}
\usepackage{xcolor}
\usepackage{spverbatim}
\usepackage{amssymb}
\usepackage{bm}
\usepackage{url}

\newcommand{\hb}[1]{\textbf{#1}}

\title{NLData2Opt: Benchmarking Parameter Derivation and Prediction for Modeling Optimization Problems from Natural Language and Data}

\author{
Lijia Liu\textsuperscript{\textnormal{1}},
Kyohei Atarashi\textsuperscript{\textnormal{1}},
Jiyi Li\textsuperscript{\textnormal{2}}, \\
\textbf{Koh Takeuchi\textsuperscript{\textnormal{1}},
Shunji Umetani\textsuperscript{\textnormal{3}},
Hisashi Kashima\textsuperscript{\textnormal{1}}}\\
\textsuperscript{1}Kyoto University \quad
\textsuperscript{2}Hokkaido University \quad
\textsuperscript{3}Recruit Co., Ltd.\\
}
\date{December 2025}

\begin{document}

\maketitle

\begin{abstract}
Formulating optimization problems from raw inputs of problem descriptions and data is time-consuming and costly. While Large Language Models (LLMs) offer a promising way to automate this process, existing benchmarks overestimate their capabilities by relying on idealized inputs where all data is relevant and complete. This diverges from reality, where data-centric reasoning is required to extract, compute, and predict parameters from noisy sources. 
To address this, we introduce NLData2Opt, a benchmark to evaluate LLMs on the multi-stage reasoning inherent in practical optimization workflows. NLData2Opt decomposes problems into problem descriptions and data, and requires models to (1) extract relevant parameters from data containing misleading attributes, (2) compute missing parameters derived by transforming existing data attributes, and (3) predict unavailable future values from historical observations. Evaluation results show substantial performance decline relative to existing benchmarks, indicating that benchmark progress does not directly translate to formulation accuracy in practical optimization workflows. NLData2Opt is available at \url{https://github.com/sumarivyn/NLData2Opt}.

\end{abstract}

\section{Introduction}

Modeling real world problems through mathematical optimization to find optimal solutions has many applications, such as task scheduling~\cite{cheng2024, pallagani2024}, vehicle routing~\cite{baty2024, luo2025}, and healthcare planning~\cite{tripathi2024, tang2024}. Recent works explore using Large Language Models (LLMs) to automatically translate natural language inputs into optimization models, making the time-consuming and costly formulation process more accessible. In this setting, the natural language input generally consists of two components that summarize the optimization problem: a problem description and its corresponding data. The problem description defines the objective and constraints, and data provides the concrete parameters required to formulate the problem. A \textit{parameter} denotes a known attribute specified in the data, while a \textit{variable} denotes a decision quantity whose value is determined by solving the optimization problem.

Current benchmarks designed to evaluate LLMs on optimization modeling generally assume clean and well-structured inputs. Specifically, they assume complete and relevant problem descriptions and data are provided. For example, NL4Opt~\cite{nl4opt}, Mamo~\cite{mamo}, OptiGuide~\cite{optiguide} and OPT-Engine~\cite{optengine} assume a single unified natural language input can be directly converted into an optimization model, and explicitly provide all parameters required for the formulation within the problem description. NLP4LP~\cite{optimus}, ComplexOR~\cite{ChainOfExperts} and ORThought~\cite{orthought} introduce problems with significantly longer natural language descriptions. Advanced benchmarks like IndustryOR~\cite{orlm} shift towards problems from different industrial sectors, while OptiVerse~\cite{optiverse} covers complex domains such as stochastic programming. Optibench~\cite{optibench} embed linear and nonlinear tabular data in the problem description to measure LLMs' end-to-end problem solving abilities.

\begin{figure*}[t]
    \centering
    \includegraphics[width=\textwidth, trim={7mm 0mm 7mm 0mm}]{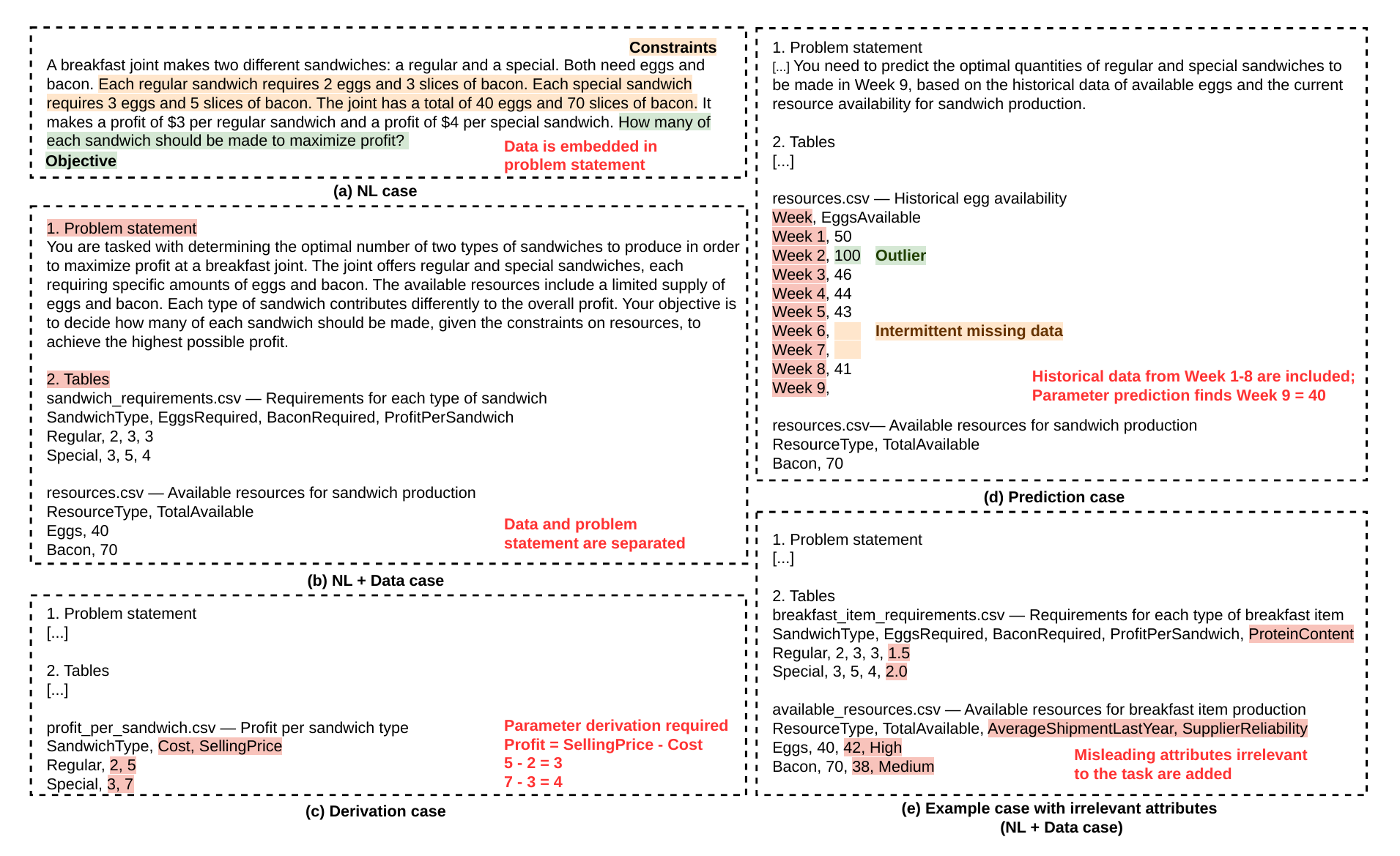}
    \caption{An optimization task in NLData2Opt with its components.}
    \label{fig:formulation}
\end{figure*}

However, despite recent progress, current benchmarks have two fundamental limitations. 
Firstly, data is directly embedded in the problem description, and the model's task is to locate and map them into an optimization formulation. This diverges from real-world modeling, in which the construction of the objective and constraints is usually a separate process from data collection, due to the scale and independent storage of the data. 
Secondly, it is assumed that every parameter is provided in the data in a ready-to-use form. This is rarely guaranteed with noisy and incomplete real-world data~\cite{bengio2021}. Since raw data typically contains several pieces of contextually related information, we have to identify through \textit{parameter extraction} what information is actually necessary to solve the problem~\cite{shi2023}.  Analysts must also frequently perform \textit{parameter derivation}, which involves transforming multiple existing data attributes into a single usable value, such as calculating net profit from revenue and expenses. In cases such as demand forecasting or production scheduling, a predict-then-optimize pipeline is commonly adopted, where \textit{parameter prediction} estimates unavailable future data from historical trends, and the predicted values are subsequently used in deterministic planning models~\cite{elmachtoub, mandi2024}.
Due to these two limitations, it is difficult to evaluate LLMs' optimization modeling capabilities in real-world applications with existing benchmarks. 

To address this, we introduce NLData2Opt, a novel benchmark that incorporates multi-stage reasoning through controlled and systematic tasks, as illustrated in Figure~\ref{fig:formulation}. Tasks are decomposed into two structurally separate components: a natural language problem description and a collection of CSV-style data tables. Noisy and incomplete data is reflected in the need to extract, compute, and predict parameters. 
Under the NLData2Opt framework, an LLM needs to fulfill three challenges to complete the optimization formulation: (1) extract relevant parameters from data containing misleading attributes, (2) compute missing parameters derived by transforming existing data attributes, and (3) predict unavailable future values from historical observations. We comprehensively evaluate state-of-the-art methods and ten open-weight and proprietary LLMs. Results show that current LLMs perform competitively in simplified tasks but achieve limited success on NLData2Opt.

\noindent\textbf{Contributions.} We make three contributions:
(1) We introduce {NLData2Opt}, a benchmark that evaluates whether LLMs can perform multi-stage reasoning for optimization formulation, including model--data separation, information retrieval, implicit derivation, and future-value prediction.
(2) We propose an {automatic construction pipeline} that transforms NLP4LP and Mamo tasks while preserving the feasible region and optimal objective value.
(3) We provide {comprehensive evaluations} of prompting, agent-based, and fine-tuned baselines, which reveal substantial gaps between structural correctness and functional executability. 

\section{Related Work}

Modeling an optimization problem into an accurate formulation is often time-consuming and requires domain expertise. To enhance this process, recent work has explored automating the formulation process, reducing the time and expertise needed. 

Chain-of-thought variants include Tree of Thoughts~\cite{tree_of_thoughts} and Autoformulation~\cite{autoformulation}, while multi-agent systems include CoE~\cite{ChainOfExperts}, OptLLM~\cite{optllm} and EoR~\cite{eor}.
Training-free approaches~\cite{optichat, cityleo, li2023} provide step-by-step instruction and incorporate prior knowledge into prompt design to improve performance.
For learning-based methods, several existing works explore applying SFT with synthetic data~\cite{orlm, wang2024, evo-step} or adopt reinforcement learning frameworks~\cite{sirl2025, foarl2025}, while LLaMoCo~\cite{llamoco} incorporates a contrastive learning-based warm-up procedure to enhance convergence behavior.

Techniques to improve formulation quality include highlighting named entities with XML-like tags~\cite{highlight_entities}. LLMOPT~\cite{llmopt} constructs a five-element formulation and reports an $11.08\%$ average improvement in solving accuracy over SOTA methods. NED-Tree~\cite{nedtree} and LEAN-LLM-OPT~\cite{leanllmopt} extend to handle complex nonlinear and large-scale industrial formulations. During inference, format control such as five-element formulation, LaTeX code, or JSON outputs~\cite{llmopt, ttg, wang2024} assists models to parse the problem into symbolic optimization models.

\section{NLData2Opt}\label{problem_def}
This section introduces NLData2Opt, a benchmark for evaluating LLMs on formulating optimization problems in realistic data conditions. We introduce the class of optimization problems considered, then outline the NLData2Opt task structures, and finally provide a validity criterion to ensure ground-truth solution validity.

\subsection{Mixed Integer Linear Program}\label{}
A mathematical optimization problem aims to find an optimal solution of decision variables that minimizes the objective function while satisfying all constraints.
In this paper, we focus on modeling Linear Program (LP) and Mixed Integer Linear Programs (MILP), which is one of the most popular optimization problems having a wide variety of industrial applications.
Given a matrix $\bm{A} \in \mathbb{R}^{m \times n}$, $\bm{b} \in \mathbb{R}_+^m$, $\bm{c} \in \mathbb{R}^n$, and a subset $I \subseteq N = \{ 1,\dots,n \}$, the corresponding MILP $\mathcal{P}$ is given by
\begin{equation}
    \begin{array}{lll}
    \textnormal{min} & \bm{c}^\top \bm{x}\\
    \textnormal{subject to} & \bm{A} \bm{x} \ge \bm{b},\\
    & x_j \ge 0, & j \in N,\\
    & x_j \in \mathbb{R}, & j \in N \setminus I,\\
    & x_j \in \mathbb{Z}, & j \in I,
    \end{array}
\end{equation}
where we call LP in case of $I = \emptyset$.
Currently, many branch-and-cut solvers have been developed for MILP, where input data is written in the standard MPS (Mathematical Programming System) or LP formats.
Alternatively, mathematical optimization problems can be specified via several modeling languages such as AMPL~\cite{ampl} and GAMS~\cite{gams}. 

\subsection{Task Definition in NLData2Opt}\label{sec:task-definition}
Given a natural language problem description $\mathcal{P}$ and a set of data $\mathcal{T}$, an LLM is required to formulate the task in AMPL. We define $\mathcal{T}=\mathcal{T}_{rel}\cup\mathcal{T}_{irr}$, where $\mathcal{T}_{rel}$ contains all parameters necessary for the optimization formulation, and $\mathcal{T}_{irr}$ denotes the attributes irrelevant to the problem. For all cases except the base case, we consider both $\mathcal{T}_{irr} = \emptyset$ and settings where $\mathcal{T}_{irr}$ contains irrelevant attributes. The objective is to produce an executable formulation that yields a solution to the optimization problem. We denote the generated AMPL model file and data file by $\mathcal{M}$ and $\mathcal{D}$, respectively.

Each instance consists of the following four standard cases, expanded into nine prompt settings.

\noindent
\textbf{NL Case (base case).}  
    The problem is expressed entirely in natural language, with all numerical parameters embedded in the text. The LLM directly generates $(\mathcal{M}, \mathcal{D})$ from $\mathcal{P}$:
    \[
        \mathcal{F}(\mathcal{P}) \rightarrow (\mathcal{M}, \mathcal{D}).
    \]

\noindent
\textbf{NL+Data Case.}  
    The problem description $\mathcal{P}$ and dataset $\mathcal{T}$ are provided separately. The LLM generates
    \[
        \mathcal{F}(\mathcal{P}, \mathcal{T}) \rightarrow (\mathcal{M}, \mathcal{D}).
    \]

\noindent
\textbf{Derivation Case.}  
    In addition to the separation of the problem description and data, each $k\in \mathbf{k}$ in a subset of parameters $\mathbf{k} \subset \mathcal{T}_{rel}$ is defined implicitly through intermediate quantities $Q_k$. The LLM must compute the parameters via an algebraic functional relationship:
    \[
        k = \phi(Q_k), \quad \forall k \in \mathbf{k},
    \]
    before constructing the optimization model. This transformation preserves the feasible region and the optimal solution while introducing derivation-oriented reasoning. The optimization problem is formulated as:
    \[
    \mathcal{F}(\mathcal{P}, (\mathcal{T} \setminus \mathbf{k}) \cup \bigcup_{k\in \mathbf{k}} Q_k) \rightarrow (\mathcal{M}, \mathcal{D}).
    \]
    
\noindent
\textbf{Prediction Case.}  
    In addition to the separation of the problem description and data, a subset of parameters $\mathbf{t} \subset \mathcal{T}_{rel}$ is masked from the data tables. Instead, for each missing parameter $t \in \mathbf{t}$, a sequence of historical observations $H_t = \{h_1, \dots, h_n\}$ ($n \ge 7$) is provided. We denote the full set of historical data as $\mathcal{H}$. The LLM must first perform a forecasting step to estimate the missing values:
\[
    \psi(\mathcal{H}) \rightarrow \hat{\mathbf{t}},
\]
where $\hat{\mathbf{t}}$ is the set of predicted values for the masked parameters. The optimization problem is formulated as:
\[
\mathcal{F}(\mathcal{P}, (\mathcal{T} \setminus \mathbf{t}) \cup  \bigcup_{t\in \mathbf{t}} H_t)) \rightarrow (\mathcal{M}, \mathcal{D}).
\]

\subsection{Validity Criterion}
To maintain the validity of the benchmark, every transformation must satisfy the following criterion. Let $S^*(\mathcal{O})$ denote the optimal objective value of an optimization problem $\mathcal{O}$. Let $\mathcal{O}_{\text{orig}}$ be the original problem and $\mathcal{O}_{\text{trans}}$ its transformed counterpart. Each transformation $\Psi \in \{\text{Base}, \text{Prediction}, \text{Derivation}\}$ is required to preserve the original feasible region and optimal solution. 
Accordingly, we ensure that
\begin{equation}
    S^*(\mathcal{O}_{\text{trans}}) = S^*(\mathcal{O}_{\text{orig}}),
\end{equation}
which guarantees that the ground-truth solution of the optimization problem remains invariant under all transformations.

\section{Automatic Construction of NLData2Opt}\label{nlp4pad}

\subsection{Separation of Problem Description and Data}\label{sec:separation}
Each task in NLData2Opt, except for the NL case, is decomposed into two structurally separate components to simulate practical optimization workflows:

\noindent
\textbf{Problem description.}  
A structured natural language description that specifies the objective and constraints, but does not include any concrete numerical values. This component guides the LLM to construct the AMPL model file.

\noindent
\textbf{Data.}  
A collection of CSV-style tables that contain numerical parameters. Each table includes a name, a brief description and multiple rows of numerical or categorical data. Provided independently of the problem description, this component guides the LLM to construct the AMPL data file.
\begin{figure*}[t]
    \centering
    \includegraphics[width=\textwidth, trim={7mm 0mm 7mm 0mm}]{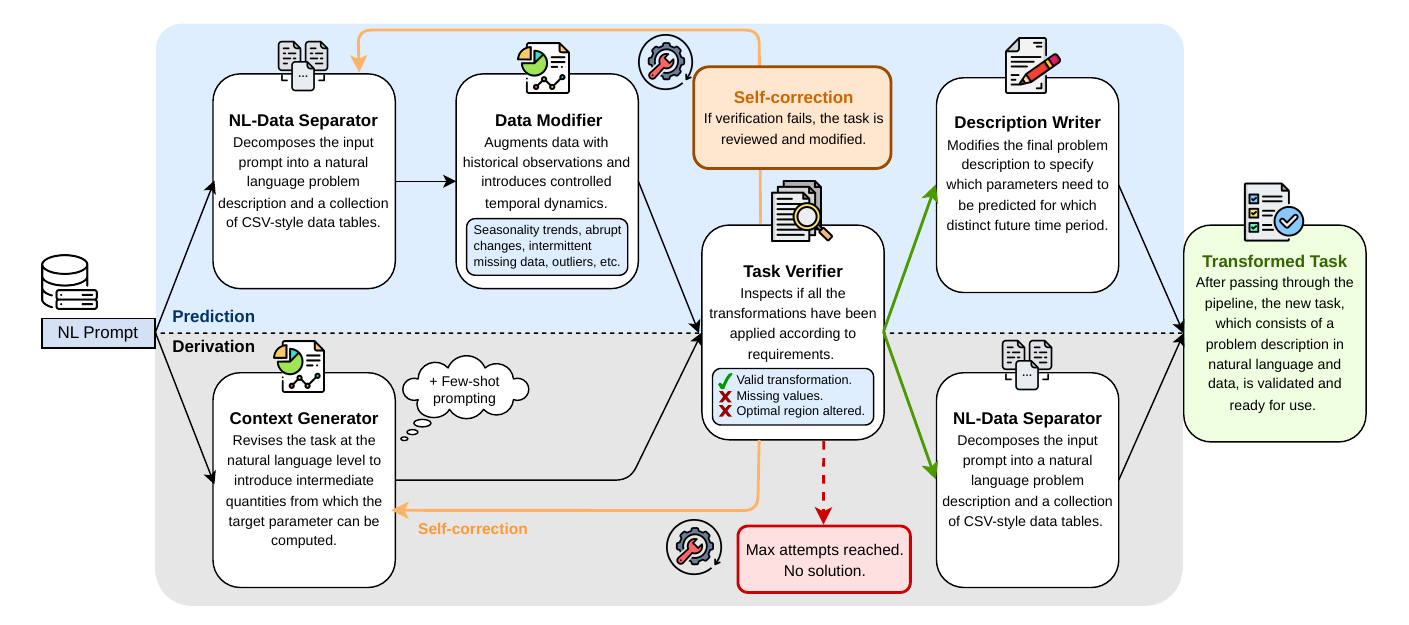}
    \caption{Transformation pipeline for prediction and derivation cases.}
    \label{fig:pipeline_pred_der}
\end{figure*}

\begin{figure*}[t]
    \centering
    \includegraphics[width=\textwidth, trim={7mm 0mm 7mm 0mm}]{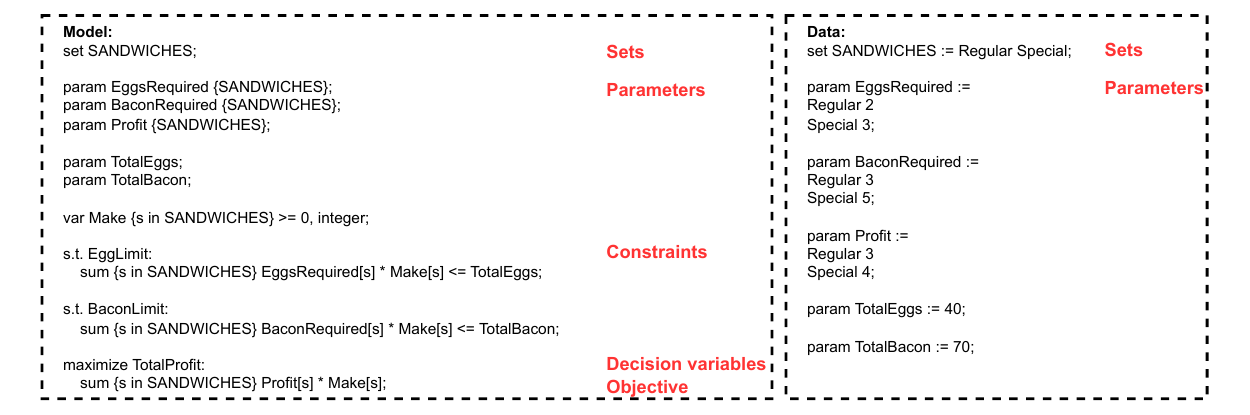}
    \caption{AMPL formulation of an example optimization task with model--data separation.}
    \label{fig:ampl_formulation}
\end{figure*}

The transformation process for the prediction and derivation cases follows a multi-stage pipeline, as illustrated in Figure~\ref{fig:pipeline_pred_der}. In the single-parameter prediction case, we first augment the data with historical observations. An AMPL Formulator samples from predefined prompt templates to produce controlled temporal patterns that systematically evaluate whether models can predict parameters and integrate them into optimization formulations. Patterns such as seasonality, abrupt changes, intermittent missing data and outliers are then introduced to simulate richer temporal dynamics.
Finally, the problem description is revised to explicitly specify the parameter to be predicted and the corresponding future time period.

The multi-parameter prediction case extends this procedure by applying the same transformation to multiple target parameters, incorporating historical data and prediction requirements for each parameter independently.

For derivation case, directly transforming existing optimization problems to require intermediate calculations in a single step proved challenging for GPT-4o. Instead, we adopt a two-stage transformation strategy. 
The Context Generator first revises the task at the natural language level to introduce intermediate quantities from which the target parameter $p$ can be computed. Only after this modification is the revised task separated into a problem description and corresponding data tables. To further encourage valid derivation-based optimization formulations, we employ few-shot prompting during generation. Detailed prompt templates are provided in Appendix~\ref{sec:pred_gen}-\ref{sec:deriv_gen}.

\subsection{Generating Gold Solutions in AMPL}\label{sec:gold-ampl}

To construct gold-standard solutions for NLData2Opt, we adopt AMPL as the modeling language for mathematical programming. This choice is motivated by several considerations. 
(1) AMPL explicitly separates the \textit{model}, which details the sets, parameters, decision variables, objectives, and constraints, from the \textit{data}, which defines the specific set and parameter values, as shown in Figure~\ref{fig:ampl_formulation}. This encourages more compact and readable model definitions, while mirroring practical scenarios where large-scale data is stored separately.
(2) Treating data as a separate component allows modifications without altering the underlying model structure, allowing the same model to be evaluated on different datasets or vice versa~\cite{ampl}. 
(3) AMPL is a declarative language based on algebraic representations of objectives and constraints, which is natural and familiar to practitioners in large-scale mathematical programming.

\begin{figure*}[t]
    \centering
    \includegraphics[width=\textwidth, trim={7mm 0mm 7mm 0mm}]{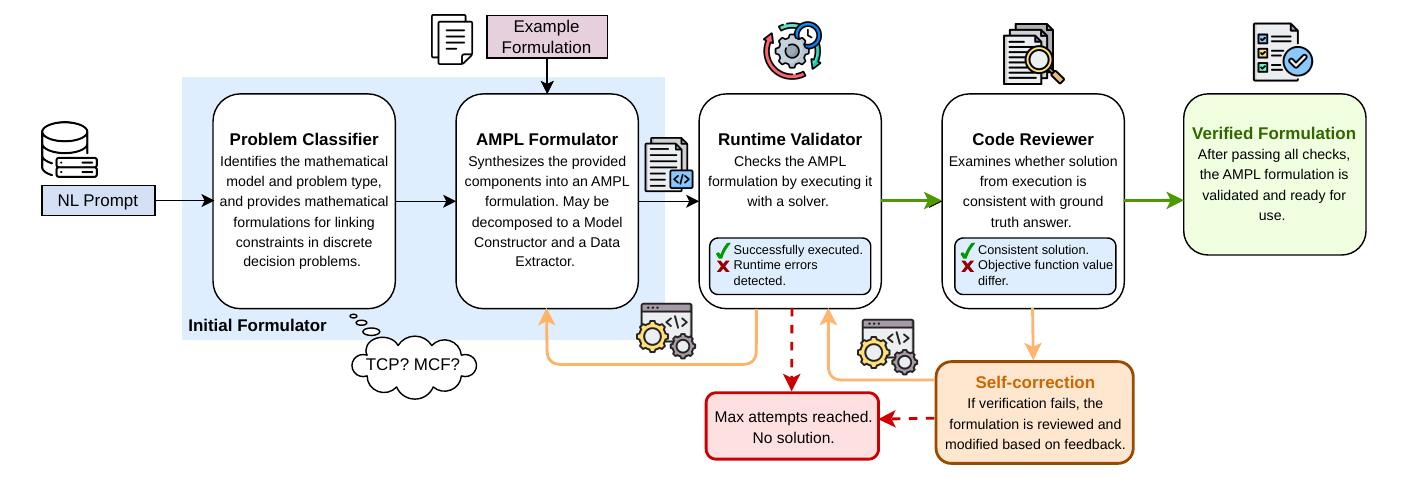}
    \caption{Ground-truth AMPL formulation construction framework.}
    \label{fig:pipeline_ampl}
\end{figure*}

To generate AMPL formulations, we adopt a CoE framework, as illustrated in Figure~\ref{fig:pipeline_ampl}. Our implementation builds on prior work~\cite{ChainOfExperts}, with prompts adapted to produce AMPL files. The system consists of multiple specialized experts, each responsible for a specific subtask. A Problem Classifier first identifies the optimization problem class (e.g., transshipment or max-flow). If a standard template is available, it is passed directly to the AMPL Formulator; otherwise, a Model Constructor and a Data Extractor generate the formulation from scratch. The resulting code is then refined through an iterative Code Runtime and Self-correction loop. Detailed prompt templates for each expert are provided in Appendix~\ref{sec:ampl_gen}.

\subsection{Evaluation Metrics}\label{sec:metrics}
Since generated formulations may be partially correct, we adopt three complementary metrics to evaluate outputs from different perspectives.

\noindent
\textbf{Entity F1 Score (EF1).}  
Structural correctness is measured using an entity-level F1 score. Each AMPL model is represented as a multiset $\mathcal{E}$ of entity types drawn from
\[
\Omega = \{\text{set}, \text{parameter}, \text{variable}, \text{constraint}, \text{objective}\}.
\]

Let $\mathcal{E}_{\text{cand}}$ and $\mathcal{E}_{\text{gt}}$ denote the multisets extracted from the candidate and ground-truth models, respectively. The number of true positives is computed as
$\mathrm{TP} = \sum_{\omega \in \Omega} \min\bigl(\mathrm{mult}(\mathcal{E}_{\text{cand}}, \omega), \mathrm{mult}(\mathcal{E}_{\text{gt}}, \omega)\bigr)$,
where $\mathrm{mult}(\mathcal{E}, \omega) := |\{ e \in \mathcal{E} \mid e = \omega \}|$ denotes the multiplicity of entity type $\omega$ in $\mathcal{E}$.
Precision and recall are defined as
$\mathrm{P} = \frac{\mathrm{TP}}{|\mathcal{E}_{\text{cand}}|}$ and 
$\mathrm{R} = \frac{\mathrm{TP}}{|\mathcal{E}_{\text{gt}}|}$, 
and the F1 score becomes
$\mathrm{F1} = \frac{2 \mathrm{P} \mathrm{R}}{\mathrm{P} + \mathrm{R}}$.
This captures structural correctness independent of variable or constraint naming.

\noindent
\textbf{Execution-Based Correctness (EBC).}  
Functional correctness is evaluated by executing each model--data pair using the AMPL interpreter with the CBC solver. If execution fails, the score is set to $0$. Otherwise, a smoothed correctness score is computed:
\begin{equation}
  \mathrm{Correctness}=
  \begin{cases} 
        \max\!\left(0, 1 - \left| \frac{O - A}{A} \right|\right) & \text{if } A \neq 0, \\
        1 & \hspace{-8mm} \text{if } O = A = 0, \\
        0 & \hspace{-1.5mm}\text{otherwise},
    \end{cases}
\end{equation}
where $O$ denotes the objective value from the candidate formulation and $A$ is the ground-truth value.

\noindent
\textbf{LLM-Based Evaluation Score (LLM-ES).}  
We  employ GPT-4o-mini to perform a holistic evaluation of candidate formulations. The evaluator compares the constraints specified in the AMPL model and the parameter values specified in the data against the corresponding components of the gold-standard formulation, assigning a score on a $0$-$10$ scale. The full evaluation prompt is provided in Appendix~\ref{sec:llm_eval}.
 
\subsection{Quality Control}\label{sec:quality-control}
We rigorously validate each generated AMPL formulation through a combination of execution-based and semantic checks. Every formulation is executed to detect runtime or modeling errors, and the resulting objective value is compared against the ground-truth solution. In addition, we employ LLM-based reviewers to perform holistic verification of formulation soundness. When validating different cases of natural language problem descriptions and data pairs, these reviewers ensure that all transformation rules are satisfied and that each transformed instance preserves the original feasible region and optimal solution.

\begin{figure*}[t]
    \centering
    \includegraphics[width=\textwidth, trim={7mm 0mm 7mm 0mm}]{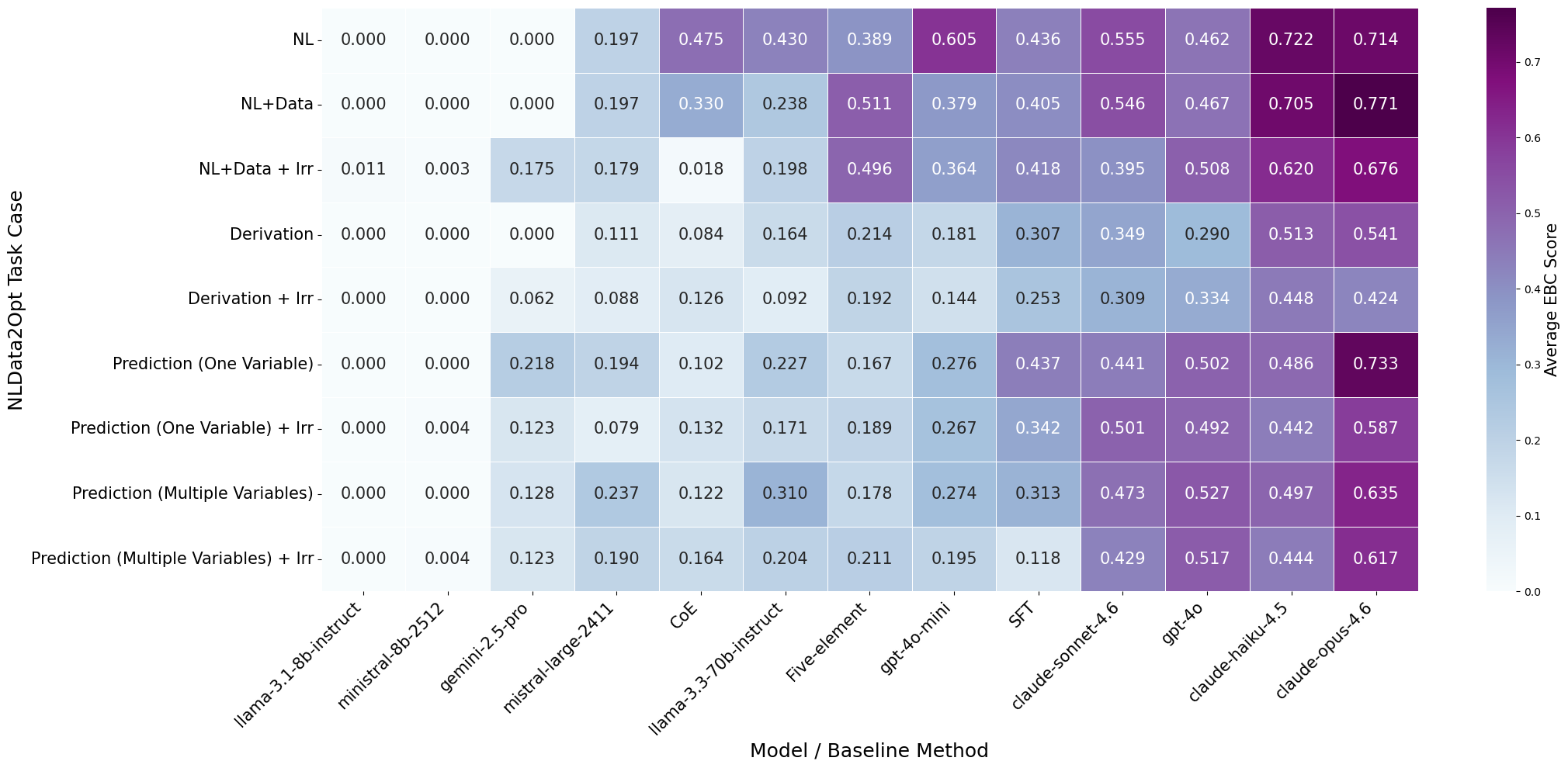}
    \caption{\textbf{Average EBC scores across task cases.} The heatmap illustrates functional correctness achieved on the nine task cases of NLData2Opt (rows, cases with irrelevant attributes are marked with "+ Irr"), when evaluated with an LLM model or baseline method (columns). The columns are ranked to reflect the worst to best performing models. Higher saturation represents a higher EBC score. While models with smaller capacity did not generate executable code, stronger models only reached scores as high as 0.77, highlighting the challenges of the tasks. }
    \label{fig:ebc_heatmap}
\end{figure*}

\section{Experiments}
We develop NLData2Opt and comprehensively evaluate SOTA methods to assess their ability to formulate and solve realistic optimization problems from natural language and data. The observed performance decline highlights the challenges posed by NLData2Opt and indicates that current methods are not yet capable of fully addressing such tasks.

\subsection{Experimental Setup}\label{sec:exp-setup}

\noindent
\textbf{Benchmarks.}  
We construct NLData2Opt from two existing benchmarks: NLP4LP~\cite{optimus} and Mamo~\cite{mamo}. NLP4LP consists of 269 LP and MILP tasks collected from optimization textbooks and lecture notes. Mamo is divided into two subsets: Easy LP, which contains 647 high school-level MILP problems, and Complex LP, which includes 210 undergraduate-level tasks integrating both LP and MILP formulations.

\noindent
\textbf{Baselines.}
We evaluate NLData2Opt using a diverse set of methods. Proprietary LLMs evaluated under direct prompting include GPT-4o-mini (\texttt{gpt-4o-mini-2024-07-18}), GPT-4o (\texttt{gpt-4o-2024-11-20}), Gemini 2.5 Pro, Claude Haiku 4.5 (\texttt{claude-haiku-4-5-20251001}), Claude Sonnet 4.6 and Claude Opus 4.6. Open-weight models include Llama-3.1-8B-Instruct, Ministral-3-8B-2512, Llama-3.3-70B-Instruct and Mistral-Large-2411. We use a temperature of $0.1$ and limit each model to a single attempt per task. Generated formulations are normalized to AMPL \texttt{.mod}/\texttt{.dat} format using the reformatting prompt in Appendix~\ref{sec:llm_eval} and executed using the AMPL interpreter with the CBC solver. Non-executable outputs are assigned zero EBC.

In addition, we evaluate several structured baselines: the five-element formulation approach proposed by LLMOPT using GPT-4o-mini, CoE with its default backbone GPT-3.5-turbo, and a SFT baseline trained on Qwen3-1.7B.

\subsection{Results and Analysis}\label{sec:results}
Figure~\ref{fig:ebc_heatmap} reports EBC across nine prompt settings: NL, NL+Data, derivation, two prediction cases, and four cases with irrelevant attributes. Comprehensive evaluation results for all metrics (EF1, EBC and LLM-ES) are summarized in Tables~\ref{tab:metrics_question}-\ref{tab:metrics_pred2_irr} in Appendix~\ref{sec:results_all}, averaged over a $15\%$ subset per dataset using a random seed of $42$. Additionally, we report full results in Tables~\ref{tab:opus_full}-\ref{tab:llama_full} for three representative models: claude-opus-4.6, gpt-4o-mini, and llama-3.3-70b-instruct.
Overall, current LLMs and other baseline methods achieve limited success on NLData2Opt, with subset-based evaluation showing trends consistent with the full benchmark.

\noindent
\textbf{Impact of model--data separation.}
Transitioning from NL (Table~\ref{tab:metrics_question}) to NL+Data cases (Table~\ref{tab:metrics_base}), LLMs struggle with mapping declared variables in the model to instantiated values in the data. Consequently, we observe a general decrease in EBC, even when EF1 remains relatively stable. 
However, we note that all proprietary models achieve higher EBC on Mamo Complex for NL+Data case. As Mamo Complex contains more complicated tasks, extracting data into separate tables might have provided more clarity and structure. 

Qualitative analysis shows that LLMs tend to embed coefficients directly into constraints in the NL case. In contrast, in the NL+Data case, LLMs use a more elegant approach by defining the parameters in the model file and recording the corresponding values in the data file. While both methods yield the same solution, parameterizing is more desirable as it allows for higher scalability and generalizability.

\noindent
\textbf{Impact of irrelevant attributes.}
Our experiments illustrate that the performance drop is not simply due to noise. Rather, when multiple conceptually related pieces of information are present, models struggle to identify the correct data. In our initial experiments, adding irrelevant attributes did not significantly affect results. However, after introducing more semantically similar attributes, we observe performance declines across all three metrics. 
Error analysis (Figure~\ref{fig:error_ampl}) further confirms this, revealing that irrelevant attributes increase undefined entities errors, including a rise from $36.3\%$ to $47.6\%$ for the derivation case. This confirms that the presence of semantically similar yet irrelevant data hinders the model's ability to maintain a mapping between declared variables and instantiated data.

\noindent
\textbf{Derivation case analysis.}
The derivation cases (Table~\ref{tab:metrics_computed}) receive drops in EBC across baselines. Table~\ref{tab:modelerror_computed} reports the percentage of correctly identified models, revealing that correct constraint structures are relatively low. This indicates that models struggle to maintain structural consistency when parameters are not explicitly provided.

Qualitative analysis further supports that models often struggle during optimization formulation that requires parameter derivation. When a total quantity must be derived from multiple components, models may fail to aggregate, generating multiple false constraints instead. Moreover, even when derivation is performed correctly, models introduce irrelevant constraints that unnecessarily restrict the feasible region. This suggests that models struggle to integrate intermediate derived results into correct downstream optimization constraints.

\noindent
\textbf{Prediction case analysis.}
Performance drops in both cases, though one-variable prediction (Table~\ref{tab:metrics_pred1}) does not show significant EBC improvement over multiple-variable cases (Table~\ref{tab:metrics_pred2}). 
Table~\ref{tab:error_pred} shows that even if an exact match is not achieved, predictions often lie in the ground truth value's vicinity.
Simple forecasting and oracle-prediction baselines (Tables~\ref{tab:opus_pred}-\ref{tab:llama_pred}) confirm that forecasting errors alone cannot explain the observed performance gap caused by formulation errors.

Qualitative analysis shows that models occasionally predict incorrect values, such as directly copying from the second-to-last time period. Others correctly predict the required parameters but fail to incorporate them into the formulation. 
Across both derivation and prediction cases, common AMPL execution errors include undefined entities and undeclared data assignments. Therefore, even when arithmetic or forecasting is correct, integrating intermediate results into downstream constraints remains a key challenge in multi-step tasks.

\subsection{Human Alignment}
Although human evaluation remains the gold standard, evaluating NLData2Opt presents scalability challenges. Each task requires the evaluation of nine distinct cases, and assessing AMPL formulations requires domain expertise. Consequently, comprehensive human evaluation on the entire benchmark across baselines is far too expensive.

To provide a scalable and cost-effective alternative, we adopt an LLM-as-a-judge framework using the GPT-4o-mini-based LLM-ES metric. To verify this approach, we manually evaluated 108 candidate formulations generated by Claude Opus 4.6. The instructions used were identical to the LLM judge's prompt (Appendix~\ref{sec:llm_eval}). The evaluation yields a Spearman’s rank correlation coefficient of $\rho = 0.39$ ($p = 3.90 \times 10^{-6}$) and a Pearson linear correlation coefficient of $r = 0.30$ ($p = 3.90 \times 10^{-4}$). These results indicate a positive alignment with human judgments, consistent with prior work on LLM-based evaluators~\cite{bavaresco2025, schroeder2024}. Results suggest that evaluating specialized programming is more challenging than evaluating open-ended natural language generation. 
While LLM-ES provides a scalable signal, its moderate correlation with human judgments suggests it should be interpreted with caution and used alongside other execution-based metrics.

\section{Conclusion}
We introduced NLData2Opt, a benchmark designed to evaluate optimization modeling under realistic data conditions. NLData2Opt reflects practical workflows by separating data from problem descriptions and incorporating irrelevant yet misleading attributes. In addition, it requires the derivation of missing parameters and prediction of future values that are common in applied optimization but largely absent from prior benchmarks. 

Through comprehensive evaluations, we demonstrate that current LLMs and baseline methods achieve only limited success on NLData2Opt. These findings underscore the need for more robust approaches to optimization problem formulation.

\clearpage
\section*{Limitations}
NLData2Opt utilizes LLM-based transformations for benchmark construction, which may contain hallucinated information, temporal inconsistencies, or other logical distortions. They may also introduce synthetic parameters that differ systematically from real-world data distributions. Common mitigation strategies include human evaluation, comparison between real and synthetic data, error analysis, and execution-based validation. In NLData2Opt, we reduce these risks through execution-based checks of generated optimization formulations and LLM-based validation of transformed instances against predefined consistency criteria. Nevertheless, automated validation cannot guarantee that all semantic or logical errors are detected, and certain distortions may remain in the benchmark.

Because NLData2Opt is derived from existing datasets such as NLP4LP and Mamo, some source instances may have appeared in the pretraining data of evaluated models. Our transformations substantially modify the original tasks, but potential data contamination cannot be fully ruled out.

Our conclusions on industrial realism and general optimization are limited to the LP/MILP task families, data perturbations, and AMPL-based formulation setting studied in this work, and should not be interpreted as directly representing the full complexity of industrial optimization problems.

Finally, while NLData2Opt requires LLMs to predict or derive parameters before generating formal optimization formulations, this workflow may emphasize formalization ability over deeper conceptual reasoning. For example, an LLM may correctly construct an AMPL formulation using a predicted parameter without fully understanding how that parameter affects the feasible region or optimal solution. Future benchmarks could evaluate conceptual understanding and decision-making independently from formulation ability, enabling more nuanced assessments of mathematical reasoning.

\section*{Ethical considerations}
NLData2Opt is constructed by adapting open-source datasets NLP4LP and Mamo. Both datasets are licensed under CC BY NC 4.0. We respect their creators’ ownership and intellectual property, and our use of these artifacts is strictly consistent with their intended purpose of evaluating the mathematical modeling capabilities of LLMs. 

NLData2Opt is a benchmarking dataset designed to evaluate LLMs on optimization tasks. The tasks presented focus solely on mathematical optimization modeling and do not contain personal or sensitive data. We plan to release NLData2Opt under the same CC BY NC 4.0 license, allowing only non-commercial use for research purposes.

AI assistants were utilized to support prompt template development and improve the clarity of this paper. We maintain full responsibility for the content, experimental design, and conclusions presented.

\section*{Acknowledgments}
This work was supported by yJST CREST Grant Number JPMJCR21D1.

\bibliography{custom}

\appendix
\renewcommand\thesection{\Roman{section}}
\section*{Appendix}
We provide the details of the exact templates utilized in the automated-generation of NLData2Opt that were omitted in the main text. GPT-4o-mini (\texttt{gpt-4o-mini-2024-07-18}) and GPT-4o (\texttt{gpt-4o-2024-11-20}) were used with a temperature of $0.1$. While GPT-4o-mini primarily constructed NL+Data and prediction cases for NLP4LP, GPT-4o was employed for more complex tasks, including derivative cases within NLP4LP and task instances that were challenging for GPT-4o-mini to process effectively.

\section{Templates for AMPL formulation construction} \label{sec:ampl_gen}
To construct high-quality and executable gold-standard solutions for NLData2Opt, we utilize a CoE framework adapted to produce AMPL code. This pipeline decomposes the complex modeling task into subtasks handled by respective experts.
The initial formulator proposed an initial AMPL formulation, and consists of the following experts:\\
\textbf{Problem Classifier}: Identifies the precise mathematical model and objective (e.g., transshipment or max-flow) while ensuring all numerical inputs are treated as generic parameters. It also provides mathematical formulations for linking constraints in discrete decision problems. If a standard template is available, it is passed directly to the AMPL Formulator.\\
\textbf{AMPL Formulator}: Synthesizes the provided components into an executable \texttt{.mod} file, adhering to the blueprint and ensuring no numerical data values are included in the model logic.\\
\textbf{Model Constructor}: Collaborates with the Data Extractor to generate the formulation from scratch if a standard template is unavailable. \\
\textbf{Data Extractor}: Extracts data from the base case and synthesizes the final components into AMPL \texttt{.mod} and \texttt{.dat} files. 

After executing the generated AMPL formulation, the following experts revise the initial formulation through an iterative loop:\\
\textbf{Debugger}: Identifies the source of numerical errors, determining if the discrepancy between predicted and expected values arises from model logic or data errors.\\
\textbf{Code Reviewer}: Conducts a detailed analysis of the previous AMPL formulation, identifying whether the discrepancy arises from model logic or data errors, and providing a corrected proposed fix.\\
\textbf{Code Modifier}: Implements the proposed modifications by the Code Reviewer to ensure the final AMPL code accurately models the optimization problem.

\begin{lstlisting}[caption={Prompt template for Problem Classifier}]
As the Optimization Modeler and Knowledge Retriever, your primary role is to create a complete, mathematically precise blueprint for the model, ensuring all necessary constraints are included. Your output MUST be generic, using parameters for all numerical inputs (costs, demands, capacities).

### 1. Model Type & Objective:
Identify the precise mathematical model (e.g., TSP, Transshipment Problem, Min-Cost Flow) and the objective (Minimize/Maximize).

### 2. Variables and Sets:
Define all necessary Sets (e.g., CITIES, I, J) and Variables (e.g., binary x[i,j], continuous flow[i,j], auxiliary u[i]).

### 3. Binary Linking Constraints Check:
For any problem involving discrete decisions (TSP, Fixed-Charge, etc.), you MUST provide the precise mathematical formulation for the **LINKING CONSTRAINTS** necessary to connect the binary variables to the continuous variables.
    * Example for Fixed-Charge: Define the constraint ensuring flow is zero if a facility is closed (using generic parameters like 's[j]' and 'y[j]').
    * Example for TSP: Define the Miller-Tucker-Zemlin (MTZ) Subtour Elimination Constraint, including all required index exclusions (e.g., i != 1 and j != 1) and using 'N' or 'card(CITIES)' for the scaling factor.

Problem Description: {q}

Output only the complete, mathematically precise list of sets, variables, objective, and constraints, strictly adhering to the instructions above, and using generic parameter names only. Do not include any specific numerical data.
"""
\end{lstlisting}

\begin{lstlisting}[caption={Prompt template for Model Formulator}]
You are a highly specialized AMPL Coder. Your sole function is to synthesize the provided components into a complete, executable AMPL model (.mod) file.

---
### 1. Model Blueprint (From Peer):
You must strictly adhere to the model type, parameters, and constraints listed here:
{constraint_output}

### 2. Synthesis Instructions:
1.  Define all sets and parameters using generic AMPL syntax (e.g., 'param cost {{SETS}};').
2.  You must not include any data assignment operators (:=) or data values in the .mod file. All data belongs in the .dat file.
3.  Define all variables, Objective function, and Constraints, integrating the complex linking constraints provided in the blueprint.

Return only the AMPL model code for the .mod file. Do not include explanations or comments anywhere.
\end{lstlisting}

\begin{lstlisting}[caption={Prompt template for Data Extractor}]
You are the final, highly specialized AMPL Coder. Your sole function is to synthesize the provided components into a complete, executable AMPL model (.mod) and data (.dat) file.
Pay attention to the units required in the final output.

The optimization problem is the following:
{q}


The model file is already attempted by your peer:
{mod}

Your job is to create the model and data files for the optimization problem.

Return both the AMPL model and data code. Do not include explanations or comments anywhere.
```
1.mod:
[.mod file content here]

2.dat:
[.dat file content here]
\end{lstlisting}

\begin{lstlisting}[caption={Prompt template for Debugger}]
As a specialized Optimization Code Reviewer, your responsibility is to identify the source of the numerical error in the provided AMPL model.
Here is the optimization problem:
```
{q}
```

Here is the previous AMPL code:
```
{ampl}
```

Return in a few short lines why the AMPL code gave the wrong answer of {predicted}, when the correct answer should be {expected}.
Propose a fix briefly.
\end{lstlisting}

\begin{lstlisting}[caption={Prompt template for Code Reviewer}]
As a specialized Optimization Code Reviewer, your responsibility is to identify the source of the numerical error in the provided AMPL model.
Your analysis must determine if the discrepancy between the PREDICTED value ({predicted}) and the EXPECTED value ({expected}) is due to a modeling error or a data error.
Now, here is the problem:
```
{q}
```

Here is the AMPL code for review:
```
{ampl}
```

Your response should follow this format:
1. Diagnosis of Error Source
[State clearly whether the error is in the Model Logic (.mod) or the Data (.dat). If the code is correct, state that the expected value is likely flawed due to source data.]

2. Code Review: Model Logic (.mod)
[Review structure, variable definitions, and constraints. Check for MTZ exclusions, capacity links, and correct Objective function.]

3. Proposed Fix
[If the error is in the model, provide the corrected AMPL code for the .mod file. If the error is in the data, specify which parameter(e.g., unit error, transcription error).]

If both the model and the data file is correct, and the error lies in the optimization problem, output 1 in the final line.
\end{lstlisting}

\begin{lstlisting}[caption={Prompt template for Code Modifier}]
As a specialized Optimization Code Reviewer, your responsibility is to improve the current AMPL model.
Here is the optimization problem:
```
{q}
```

Here is the previous AMPL code:
```
{ampl}
```

Here is the comment and proposed fix by your peer:
```
{comment}
```

Modify the AMPL code such that it models the question accurately, based on the comment and proposed fix by your peer.
Return only the new AMPL code. Do not include explanations or comments anywhere.
Use the exact format below:
```
1.mod:
[.mod file content here]

2.dat:
[.dat file content here]
```
\end{lstlisting}

\section{Templates for model--data separation} \label{sec:model-data}
To systematically separate the natural language description from its associated data, we employ a pipeline consisting of a NL-Data Separator and a Verifier.\\
\textbf{Model-Data Separator}: Transforms a given optimization problem into two distinct components: a problem statement in natural language and a dataset in CSV format. The problem statement presents a structured task description and specifies the optimization objective without including any numerical values. All quantitative information is captured exclusively in the dataset, which contains only the values explicitly provided in the original problem.\\
\textbf{Verifier}: Ensures the integrity of the model--data separation by checking that all numerical values required to solve the optimization problem are present in the data file.

\begin{lstlisting}[caption={Prompt template for NL-Data Separator}]
The following is an optimization problem. Please construct:
1. A problem statement in natural language, beginning with "You are..." and following a structured task description. It should clearly explain what the objective is, but must NOT contain any specific numerical values. No concrete quantities, figures, or units of measurement should be mentioned.

2. A dataset composed of multiple CSV-style tables that define a specific parameters.
   - All variables referenced in the optimization problem must be represented in the tables.
   - All table cells should be numeric, do not use placeholders.
   - Only include values explicitly mentioned in the problem.

The optimization problem is as follows:
{ques}

Please construct a problem specification and tabular dataset according to the above requirements, and provide the results the following format:
```
1. Problem statement
  [Write the problem specification in plain English.]

2. Tables
table1_name.csv - Description
Column1, Column2, ...
value1, value2, ...
...

table2_name.csv - Description
Column1, Column2, ...
value1, value2, ...
...

```
\end{lstlisting}

\begin{lstlisting}[caption={Prompt template for Verifier}]
You are asked to compare an optimization problem and its corresponding tabular datasets.
The optimization problem is as follows:
{ques}

The tabular datasets corresponding are as follows:
{dat}

Please check if the tabular datasets contain all values needed to solve the optimization problem.
If this is satisfied, return 1. If not, return 0. Return 1 or 0 only.
\end{lstlisting}

\section{Templates for prediction case construction} \label{sec:pred_gen}
To construct prediction cases for NLData2Opt, we transform NL+Data cases into tasks that require LLMs to predict future parameter values from historical data. To construct tasks that require different numbers of parameters to be predicted, we implement a pipeline that consists of the following components:\\
\textbf{AMPL Formulator}: Modifies the data of a task from NL+Data case to create prediction cases. It handles single or multiple targets, and supports increasing and stable trends. Historical data are generated for each target, with the final time period containing the true value. After passing the Verifier, this value is subsequently blanked out in the historical data table using automated scripts. Based on these patterns, seasonality, abrupt changes, intermittent missing data, and outliers are introduced to simulate richer temporal dynamics. \\
\textbf{Verifier}: Inspects the new optimization problem if all data in the NL+Data case are included. In addition, historical data are provided for a certain number of parameters, with the final time period containing the value to be predicted. Two separate verifiers are used for single variable and multi-variable prediction, respectively. \\
\textbf{Description Writer}: Generates a natural language problem description from the tabular datasets. In addition to clearly stating the overall optimization objective, it explicitly specifies which variables must be predicted at which exact time period.

\begin{lstlisting}[caption={Prompt template for AMPL Formulator (increasing case, single variable)}]
The following are a problem description and its data tables:
```
{base}
```

Modify the tabular datasets following the below requirements:
- Choose one value in the tabular datasets of the optimization problem as the prediction target.
- Provide historical data of at least 8 time periods that show a realistic trend with small random noise.
- The historical data show an increasing trend that approaches the true target value naturally, without overshooting it.
- The historical trend must stop immediately at the true target value.
- The time period should have its own column, with realistic names and values.
- Do not remove or rename any existing columns or rows in the original tables. However, the single true target value must be visibly blanked out exactly where it originally appeared.

Provide the results the following format, do not add comments:
```
1. Problem statement
  [The original problem statement]

2. Tables
table1_name.csv - Description
Column1, Column2, ...
value1, value2, ...
...

table2_name.csv - Description
Column1, Column2, ...
value1, value2, ...
...

```
\end{lstlisting}

\begin{lstlisting}[caption={Prompt template for AMPL Formulator (stable case, multiple variables)}]
The following are a problem description and its data tables:
```
{base}
```

Modify the tabular datasets following the below requirements:
- The given problem description and tables already contain one value that has been blanked out as an existing prediction target. Keep that historical data table unchanged.
- Select one additional value in the tabular datasets to serve as a second new prediction target. Blank it out exactly where it originally appeared, without removing or renaming any columns or rows.
- For the new prediction target, provide at least 8 time periods of historical data in a separate table.
- The historical series for the new target must show natural small random fluctuations around the prediction target while maintaining integer constraints when necessary.
- The historical data should hover both slightly above and below the new prediction target, centering roughly on it.
- The final row for all historical data should have the target value exactly, do not leave it blank.
- The time period should have its own column, with realistic names and values.
- Do not remove or rename any existing data aside from blanking out the two true target values.

The final sentence in the problem description should specify which two parameters need to be predicted at which time period. Phrase it as "you need to predict...".

Provide the results the following format, do not add comments:
```
1. Problem statement
  [Rewrite the original problem statement]

2. Tables
table1_name.csv - Description
Column1, Column2, ...
value1, value2, ...
...

table2_name.csv - Description
Column1, Column2, ...
value1, value2, ...
...

```
\end{lstlisting}

\begin{lstlisting}[caption={Prompt template for Verifier (single variable)}]
You are asked to verify if the given tabular datasets satisfy certain rules.

Optimization problem:
{ques}

Tabular datasets:
```
{dat}
```

Check if the data tables satisfy the following rules:
1. Historical data is only provided for 1 variable in the data tables.
2. All variables mentioned in the optimization problem appear in the main data tables except for one missing value.
3. A historical table provides the correct value for one variable, with the value in the final time period appearing in the question.

Output "1" if all rules are satisfied, otherwise output "0". Do not provide any extra text.
\end{lstlisting}

\begin{lstlisting}[caption={Prompt template for Verifier (multiple variable)}]
You are asked to verify if the given tabular datasets satisfy certain rules.

Optimization problem:
{ques}

Tabular datasets:
```
{dat}
```

Check if the data tables satisfy the following rules:
1. Historical data ending at the same numerical value as the optimization problem are provided for at least 2 variables. Ignore textual differences when comparing numerical values.
2. All variables mentioned in the optimization problem appear in the main data tables except for those in the historical tables.

Output "1" if all rules are satisfied, otherwise output "0". Do not provide any extra text.
\end{lstlisting}

\begin{lstlisting}[caption={Prompt template for Description Writer}]
The following is an optimization problem, with its corresponding tabular datasets. You are asked to write a problem description.

Optimization problem:
{ques}

Tabular datasets:
{dat}

Please write a problem description in natural language, beginning with "You are..." and following a structured task description.
Do not include any specific numerical values. No concrete quantities, figures, or units of measurement should be mentioned.
It should clearly explain what the overall objective of the optimization problem is.
Specify which parameters needs to be predicted at which time period. Phrase it as "you need to predict...". Explicitly mention the **exact time** to be predicted as it appears of the provided tables. Do not paraphrase it as "the final row" or "the next time".

Provide only the problem statement, do not add comments.
\end{lstlisting}

\section{Templates for derivation case construction} \label{sec:deriv_gen}
To generate derivation cases for NLData2Opt, we transform straightforward optimization problems into new problems that require intermediate calculations. Because directly transforming existing optimization problems to require intermediate calculations in a single step proved challenging for LLMs, we adopt a two-stage transformation strategy. The components of the pipeline are as follows:\\
\textbf{Context Generator}: Produces a rewritten optimization problem where exactly one numerical quantity is expressed via meaningful intermediate quantities, without changing the total value or feasible region. Example transformations are provided to encourage the LLM to generate valid decompositions. Example transformations are provided to guide the LLM toward valid decompositions. Two context generators are used sequentially, with context generator $2$ employed after context generator $1$ to process questions that were not successfully transformed according to the Verifier.\\
\textbf{Verifier}: Inspects the new optimization problem if it maintains the same feasible region and optimal solution as the problem before the transformation. In addition, it makes sure the transformation successfully introduced intermediate quantities to require derivation-based manipulation by making sure one numerical quantity is decomposed into possible components. \\
\textbf{NL-Data Separator}: Separates the model from its associated data for the new optimization problem. Compared to the standard NL+Data Separator used for the NL+Data case, this version specifies to only include the intermediate quantities when both aggregated parameters and intermediate quantities are provided, ensuring that derivation-based reasoning is required while modeling the optimization problem. 

\begin{lstlisting}[label={lst:deriv_context1}, caption={Prompt template for Context Generator 1}]
Rewrite the given optimization problem so that exactly one numerical variable is optionally expressed in terms of meaningful intermediate components, without changing the total value or the feasible region of the problem. All other parts of the problem must remain unchanged. The decomposition should reflect derivation-oriented reasoning, such as computing profit from revenue and costs, or breaking down total production time into separate tasks. Return only the rewritten problem.
---
The optimization problem is as follows:
A school needs to provide lunch for at least 300 students.

Rewrite:
A school needs to provide lunch for at least 150 boys and 150 girls.
---
The optimization problem is as follows:
It makes a profit of $3 per regular sandwich and a profit of $4 per special sandwich.

Rewrite:
A regular sandwich costs $1 to make and has a selling price of $4. A special sandwich costs $3 to make and has a selling price of $7.
---
The optimization problem is as follows:
{ques}

Rewrite:
\end{lstlisting}

\begin{lstlisting}[label={lst:deriv_context2}, caption={Prompt template for Context Generator 2}]
You are an Optimization Problem Decomposer. Your task is to take a straightforward optimization problem and rewrite it into a new, more detailed narrative.

You must follow these rules:

1. You *must* take the simple numbers (constants) in the problem (e.g., 10, 6, 150) and break them down arithmetically into nontrivial new components.
* FAILURE: A trivial split like '100' -> '50+50' is a **complete failure**.
* GOOD EXAMPLE: A good split adds new concepts: '4' -> 'a 1-unit prep fee and 3 units of application time'.

2. You must rewrite *only* the parts of the sentence that contain the numbers you are decomposing. **Do not** add extra, unnecessary sentences or rephrase parts of the problem that are not being changed.

3. Simply rephrasing the problem in new words or translating it into its algebraic form (like "minimize 10x + 6y") is a **complete failure**. The rewrite *must* contain the arithmetic decomposition from Rule 1.

4. The final, rewritten problem must be *mathematically identical* to the original. The objective function and all constraints must simplify to the *exact same equations*.

5. The rewritten problem should preserve the narrative style, sentence structure, and context around the numbers. It should read like a natural problem statement, not a formula or explanation.

---
**Example 1: Objective Function**
The optimization problem is as follows:
A company earns a profit of $5 per product.

Rewrite:
Each product sells for $15 and costs $10 to produce.

---
**Example 2: Simple Constraint**
The optimization problem is as follows:
A factory produces 800 units per day.

Rewrite:
The factory operates for 10 hours a day. It spends 2 hours setting up machines and then produces 100 units per hour for the remaining 8 hours.

---
**Example 3: Constraint with Multipliers**
The optimization problem is as follows:
Four times the man-hours spent on task X plus those spent on task Y should not exceed 16

Rewrite:
Each unit of task X contributes 4 man-hours, consisting of 2 hours of planning and 2 hours of execution; each unit of task Y contributes 1 man-hour of work. Then, summing these contributions, the total man-hours 4X + Y cannot exceed 16.

---
**TASK**
The optimization problem is as follows:
{ques}

Rewrite:
\end{lstlisting}

\begin{lstlisting}[label={lst:deriv_verifier}, caption={Prompt template for Verifier}]
Compare the two optimization problems below and determine if the rewritten one correctly follows the transformation rules:

1. The rewritten problem must have the same feasible region and optimal solution as the original problem after the transformation.
2. At least one numerical quantity from the original problem is split into meaningful components.

Original problem:
{ques}

Rewritten problem:
{new}

Return 1 if both rules are satisfied, else 0. Only return 1 or 0.
If 0, in the next paragraph, change the rewritten problem to fulfill the rules.
\end{lstlisting}

\begin{lstlisting}[label={lst:deriv_separator}, caption={Prompt template for NL-Data Separator}]
The following is an optimization problem. Please construct:
1. A problem statement in natural language
   - Begin with "You are..." and following a structured task description.
   - Include only those constraints that are not already evident or numerically encoded in the dataset tables.
   - Do not include any numbers, units, or placeholders anywhere in the text.

2. A dataset composed of multiple CSV-style tables that define a specific parameters.
   - All variables referenced in the optimization problem must be represented in the tables.
   - All table cells should be numeric, do not use placeholders.
   - Only include values explicitly mentioned in the problem.
   - When both aggregated parameters and intemediate parameters are provided, only include the intermediate parameters.

The optimization problem is as follows:
{ques}

Please construct a problem specification and tabular dataset according to the above requirements, and provide the results the following format:
```
1. Problem statement
  [Write the problem specification in plain English.]

2. Tables
table1_name.csv - Description
Column1, Column2, ...
value1, value2, ...
...

table2_name.csv - Description
Column1, Column2, ...
value1, value2, ...
...

```
\end{lstlisting}

\section{Templates for LLM evaluation} \label{sec:llm_eval}
In this section, instructions templates to assess the accuracy of AMPL formulation are introduced. 
Given that LLM outputs may contain conversational headers or omit the target \texttt{.mod}/\texttt{.dat} markers, we utilize the template detailed in Listing~\ref{lst:eval_reformat} to extract the AMPL formulation and ensure the files are in a compatible format with the CBC solver.

To perform a holistic evaluation, in addition to the EF1 and EBC, we employ LLM-based evaluation with the template provided in Listing~\ref{lst:eval_eval}. The rating of $0$-$10$ is then normalized to a score between $0$ and $10$, which is consistent with the other two metrics.

\begin{lstlisting}[label={lst:eval_reformat}, caption={Prompt template for reformatting generated messy formulation}]
You are given an output. Extract the AMPL formulation if possible. For the data file, leave #empty if empty, but do not omit the "2.dat" marker.

Output:
{AMPL_Formulation}

Return strictly in this format:
```1.mod:
<AMPL model>

2.dat:
<AMPL data>```

If the text does not contain tangible AMPL formulation, return:
``` 1.mod:
#empty

2.dat:
#empty ```
\end{lstlisting}

\begin{lstlisting}[label={lst:eval_eval}, caption={Prompt template for evaluating overall correctness}]
Does this .mod and .dat pair represent the same optimization problem as the ground-truth .mod and .dat? Rate this solution from 0 to 10:
.mod file:
{mod1_content}
ground truth .mod file:
{mod2_content}

.dat file:
{dat1_content}
ground truth .dat file:
{dat2_content}

Specifically, evaluate:
- Whether the constraints in the candidate .mod file are equivalent to those in the ground-truth .mod file,
- Whether the parameter values in the candidate .dat file exactly match those in the ground-truth .dat file.
Return an explanation. The final line should be the rating.
\end{lstlisting}

\section{Controlled Prediction Baselines} \label{sec:pred_analysis}
We implement two additional controlled baselines to separate forecasting difficulty from formulation difficulty, focusing on three representative models: claude-opus-4.6 (Table~\ref{tab:opus_pred}), gpt-4o-mini (Table~\ref{tab:gpt4omini_pred}), and llama-3.3-70b-instruct (Table~\ref{tab:llama_pred}). In the simple forecasting baseline, the LLM-generated predictions are replaced with estimates from a linear trend extrapolation over historical series data, while the remaining formulation process remains unchanged. In the oracle-prediction baseline, the predicted values are replaced with the ground-truth future values, thereby removing forecasting error and allowing us to examine errors arising from incorporating these values into the final optimization formulation. 

We observe that replacing the LLM-generated predictions with the simple forecasting baseline generally reduces EBC. However, using oracle-prediction values yields only small improvements over the original LLM baseline. This suggests that the forecasting capability of modern LLMs is already sufficient for many tasks and that forecasting errors alone cannot explain the observed performance gap.

The oracle-prediction baseline generally improves EBC and LLM-ES only slightly. For example, EBC of Llama 3.3 70B Instruct on the One-Variable Prediction case increases from 0.1824 to just 0.1847 under oracle prediction. This indicates that formulation errors remain persistent even when forecasting difficulty is isolated. In particular, lower-capacity models continue to struggle with incorporating predicted values into the optimization instance and constructing the corresponding formulation correctly.

\begin{table*}[h]
    \centering
    \footnotesize
    \setlength{\tabcolsep}{3pt}
    \begin{tabular*}{\textwidth}{@{\extracolsep{\fill}}l | ccc | ccc | ccc@{}}
        \toprule
        & \multicolumn{3}{c|}{Simple Forecasting} & \multicolumn{3}{c|}{Oracle Prediction} & \multicolumn{3}{c}{LLM} \\
        Case & EF1 & EBC & LLM-ES & EF1 & EBC & LLM-ES & EF1 & EBC & LLM-ES \\
        \midrule
        \quad One-Variable                & 0.8665 & 0.6144 & 0.5377 & 0.8654 & 0.7305 & 0.5858 & 0.8836 & 0.7357 & 0.5572 \\
        \quad One-Variable + Irr          & 0.8551 & 0.4796 & 0.4655 & 0.8436 & 0.6538 & 0.5018 & 0.8441 & 0.5331 & 0.4602 \\
        \quad Multi-Variable            & 0.8739 & 0.5810 & 0.5552 & 0.8643 & 0.6257 & 0.5886 & 0.8626 & 0.6225 & 0.5300 \\
        \quad Multi-Variable + Irr      & 0.8424 & 0.4972 & 0.4482 & 0.8325 & 0.5898 & 0.4769 & 0.8683 & 0.6393 & 0.4671 \\
        \bottomrule
    \end{tabular*}
    \caption{Performance breakdown for Claude Opus 4.6 across prediction baselines (Simple Forecasting, Oracle Prediction, and LLM).}
    \label{tab:opus_pred}
\end{table*}

\begin{table*}[h]
    \centering
    \footnotesize
    \setlength{\tabcolsep}{3pt}
    \begin{tabular*}{\textwidth}{@{\extracolsep{\fill}}l | ccc | ccc | ccc@{}}
        \toprule
        & \multicolumn{3}{c|}{Simple Forecasting} & \multicolumn{3}{c|}{Oracle Prediction} & \multicolumn{3}{c}{LLM} \\
        Case & EF1 & EBC & LLM-ES & EF1 & EBC & LLM-ES & EF1 & EBC & LLM-ES \\
        \midrule
        \quad One-Variable                & 0.8968 & 0.2213 & 0.5390 & 0.8548 & 0.2224 & 0.6043 & 0.3788 & 0.2386 & 0.5368 \\
        \quad One-Variable + Irr          & 0.9208 & 0.1671 & 0.4878 & 0.8765 & 0.1945 & 0.4716 & 0.4259 & 0.2555 & 0.4443 \\
        \quad Multi-Variable            & 0.9535 & 0.2229 & 0.5239 & 0.8582 & 0.2798 & 0.5830 & 0.3837 & 0.2680 & 0.5522 \\
        \quad Multi-Variable + Irr      & 0.8714 & 0.1512 & 0.4692 & 0.8500 & 0.1533 & 0.4833 & 0.3419 & 0.2071 & 0.4434 \\
        \bottomrule
    \end{tabular*}
    \caption{Performance breakdown for GPT-4o Mini across prediction baselines (Simple Forecasting, Oracle Prediction, and LLM).}
    \label{tab:gpt4omini_pred}
\end{table*}

\begin{table*}[h]
    \centering
    \footnotesize
    \setlength{\tabcolsep}{3pt}
    \begin{tabular*}{\textwidth}{@{\extracolsep{\fill}}l | ccc | ccc | ccc@{}}
        \toprule
        & \multicolumn{3}{c|}{Simple Forecasting} & \multicolumn{3}{c|}{Oracle Prediction} & \multicolumn{3}{c}{LLM} \\
        Case & EF1 & EBC & LLM-ES & EF1 & EBC & LLM-ES & EF1 & EBC & LLM-ES \\
        \midrule
        \quad One-Variable                & 0.8756 & 0.0962 & 0.4874 & 0.8007 & 0.1847 & 0.5626 & 0.3486 & 0.1824 & 0.4932 \\
        \quad One-Variable + Irr          & 0.8004 & 0.0733 & 0.4308 & 0.8046 & 0.0999 & 0.4493 & 0.1557 & 0.1180 & 0.4176 \\
        \quad Multi-Variable            & 0.8087 & 0.1387 & 0.5107 & 0.8681 & 0.2397 & 0.5739 & 0.4154 & 0.2825 & 0.5045 \\
        \quad Multi-Variable + Irr      & 0.6693 & 0.1066 & 0.4305 & 0.7159 & 0.1425 & 0.4543 & 0.2925 & 0.2008 & 0.3960 \\
        \bottomrule
    \end{tabular*}
    \caption{Performance breakdown for Meta LLaMA 3.3 70B Instruct across prediction baselines (Simple Forecasting, Oracle Prediction, and LLM).}
    \label{tab:llama_pred}
\end{table*}

\section{Further Analysis} \label{sec:further_analysis}
Table~\ref{tab:modelerror_computed} summarizes the percentage of tasks where the optimization model is correctly identified for the derivation case. Here, a correct model is when the constraint relation types ($\le$, $\ge$, $=$) and objective function type (maximize or minimize) exactly match the ground truth formulation, regardless of the values of the parameters.

\begin{table}[h]
    \centering
    \footnotesize
    \setlength\tabcolsep{2pt}
    \renewcommand{\arraystretch}{0.9}
    \begin{tabularx}{\linewidth}{X w{c}{4em} w{c}{4em} w{c}{4em}}
        \toprule
        \hb{Model/Method} & \hb{NLP4LP} & \hb{Mamo E.} & \hb{Mamo C.} \\
        \midrule
        GPT-4o-mini & 0.000 & 0.000 & 0.000 \\
        GPT-4o & 0.425 & 0.367 & 0.467 \\
        Gemini-2.5-Pro & 0.000 & 0.067 & 0.000 \\
        Claude Haiku 4.5 & 0.600 & 0.587 & 0.593 \\
        Claude Sonnet 4.6 & 0.573 & 0.533 & 0.307 \\
        Claude Opus 4.6 & 0.567 & 0.547 & 0.440 \\
        \midrule
        Llama-3.1-8B-Instruct & 0.250 & 0.165 & 0.129 \\
        Ministral-3-8B-2512 & 0.125 & 0.052 & 0.097 \\
        Llama-3.3-70B-Instruct & 0.250 & 0.300 & 0.129 \\
        Mistral-Large-2411 & 0.300 & 0.067 & 0.194 \\
        \midrule
        Five-element formulation & 0.525 & 0.794 & 0.419 \\
        CoE & 0.450 & 0.309 & 0.129 \\
        SFT & 0.375 & 0.845 & 0.355 \\
        \bottomrule
    \end{tabularx}
    \caption{Percentage of correctly identified optimization models on derivation case.}
    \label{tab:modelerror_computed}
\end{table}

Table~\ref{tab:error_pred} analyzes the performance of the prediction step prior to optimization. It reports percentage of tasks where future prediction exactly matches the ground truth value (EC) and relative correctness (RC). We find that most methods obtain over $30\%$ on the EC metric and over $60\%$ on the RC metric.

\begin{table*}[h]
    \centering
    \footnotesize
    \setlength{\tabcolsep}{2pt}
    \begin{tabular*}{\textwidth}{@{\extracolsep{\fill}}l | cc  cc | cc  cc@{}}
        \toprule
        & \multicolumn{4}{c|}{\textbf{One-variable}} 
        & \multicolumn{4}{c}{\textbf{Multiple-variables}} \\
        & \multicolumn{2}{c}{\textbf{Standard}} & \multicolumn{2}{c|}{\textbf{Irr. attributes}}
        & \multicolumn{2}{c}{\textbf{Standard}} & \multicolumn{2}{c}{\textbf{Irr. attributes}} \\
         \textbf{Model/Method} & EM & SC & EM & SC & EM & SC & EM & SC \\
        \midrule
        \quad GPT-4o Mini & 0.433 & 0.779 & 0.322 & 0.711 & 0.371 & 0.526 & 0.287 & 0.506 \\
        \quad GPT-4o & 0.611 & 0.892 & 0.400 & 0.845 & 0.408 & 0.560 & 0.258 & 0.511 \\
        \quad Gemini 2.5 Pro & 0.078 & 0.219 & 0.022 & 0.125 & 0.102 & 0.146 & 0.069 & 0.107 \\
        \quad Claude Haiku 4.5 & 0.556 & 0.972 & 0.478 & 0.961 & \textbf{0.603} & 0.744 & 0.521 & 0.718 \\
        \quad Claude Sonnet 4.6 & \textbf{0.667} & \textbf{0.980} & \textbf{0.567} & 0.942 & 0.590 & \textbf{0.745} & \textbf{0.600} & \textbf{0.745} \\
        \quad Claude Opus 4.6 & 0.533 & 0.959 & 0.511 & \textbf{0.958} & 0.571 & 0.739 & 0.563 & 0.738 \\
        \midrule
        \quad LLaMA 3.1 8B Inst. & 0.344 & 0.615 & 0.400 & 0.620 & 0.330 & 0.458 & 0.217 & 0.386 \\
        \quad Ministral 8B & 0.278 & 0.441 & 0.189 & 0.411 & 0.260 & 0.388 & 0.150 & 0.287 \\
        \quad LLaMA 3.3 70B Inst. & 0.422 & 0.846 & 0.333 & 0.820 & 0.335 & 0.522 & 0.196 & 0.452 \\
        \quad Mistral Large & 0.522 & 0.894 & 0.411 & 0.813 & 0.409 & 0.571 & 0.280 & 0.501 \\
        \midrule
        \quad Five-element & 0.389 & 0.661 & 0.411 & 0.752 & 0.327 & 0.503 & 0.303 & 0.491 \\
        \quad CoE & 0.367 & 0.638 & 0.356 & 0.700 & 0.413 & 0.508 & 0.284 & 0.435 \\
        \quad SFT & 0.544 & 0.680 & 0.533 & 0.749 & 0.469 & 0.547 & 0.411 & 0.497 \\
        \bottomrule
    \end{tabular*}
    \caption{Percentage of Exact Matches (EM) and continuous Smoothed Correctness (SC) across four prediction cases.}
    \label{tab:error_pred}
\end{table*}

\begin{figure*}[t]
    \centering
    \includegraphics[width=\textwidth, trim={7mm 0mm 7mm 0mm}]{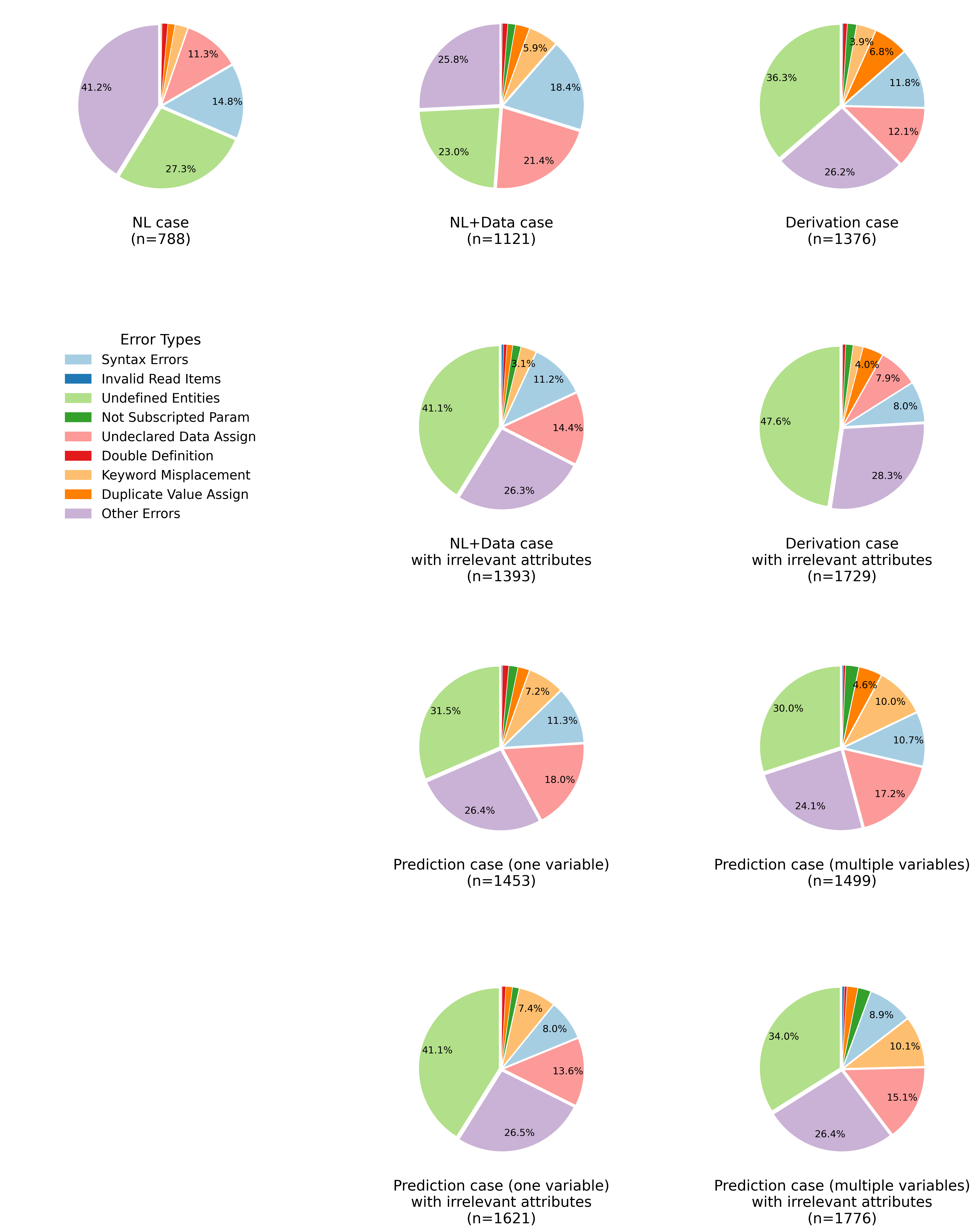}
    \caption{\textbf{Distribution of AMPL execution errors across nine task cases.} The breakdown highlights the relative frequency of specific compilation and logical errors, such as undefined entities, undeclared data assignments, and syntax errors.}
    \label{fig:error_ampl}
\end{figure*}

\begin{table*}[h]
    \centering
    \footnotesize
    \setlength{\tabcolsep}{6pt}
    \begin{tabular*}{\textwidth}{@{\extracolsep{\fill}}l | cccc@{}}
        \toprule
        \textbf{Case} & \textbf{Compilation Success} & \textbf{Exact Match Rate} & \textbf{Syntax Error} & \textbf{Structural Error} \\
        \midrule
        \quad NL                             & 51\% & 23\% & 11\% & 43\% \\
        \quad NL + Data                      & 45\% & 22\% & 24\% & 46\% \\
        \quad NL + Data + Irr                & 48\% & 25\% & 16\% & 46\% \\
        \midrule
        \quad Derivation                     & 42\% &  8\% & 20\% & 50\% \\
        \quad Derivation + Irr               & 43\% & 10\% & 12\% & 49\% \\
        \midrule
        \quad Prediction (1-Var)             & 37\% &  7\% & 17\% & 54\% \\
        \quad Prediction (1-Var) + Irr       & 35\% & 12\% &  8\% & 63\% \\
        \quad Prediction (Multi-Var)         & 41\% &  8\% & 16\% & 52\% \\
        \quad Prediction (Multi-Var) + Irr   & 38\% & 11\% & 13\% & 58\% \\
        \bottomrule
    \end{tabular*}
    \caption{Failure type breakdown across all task cases.}
    \label{tab:failure_breakdown}
\end{table*}

To further investigate the failure modes of the generated optimization models, we categorize AMPL execution errors encountered across all models and datasets. Figure~\ref{fig:error_ampl} visualizes the proportional distribution of the error types across the nine distinct task cases, while Table~\ref{tab:failure_breakdown} reorganizes failures into tool-specific syntax errors and structural errors. 

Across task cases, maintaining structural consistency between the model and the data is demonstrated to be a significant challenge for LLMs. We identify three primary types of errors that are prevalent across tasks:
\begin{itemize}
\item \textbf{Undefined Entities:} A structural error where the model (\texttt{.mod}) attempts to use a set, parameter, or variable in a constraint or objective function that has not been formally declared or initialized.
\item \textbf{Undeclared Data Assignments:} A structural error where the data file (\texttt{.dat}) attempts to assign values for parameters or sets that do not exist in the corresponding \texttt{.mod} file.
\item \textbf{Syntax Errors:} Fundamental formatting violations, such as missing semicolons or incorrect assignment operators, prevent the AMPL compiler from parsing the scripts. This includes dimension mismatches such as improper set index, reflecting a failure of mathematical indexing as well as grammar.
\end{itemize}

Beyond compile-time failures, semantic modeling errors cannot be caught during compilation but manifest as discrepancies in the final objective value. As shown in Table~\ref{tab:failure_breakdown}, the exact objective match rate is consistently lower than the compilation success rate, which indicates that many generated formulations are executable but still contain semantic modeling errors that lead to incorrect solutions.

\section{Comprehensive Evaluation Results} \label{sec:results_all}
The experimental results reported in Tables~\ref{tab:metrics_question}-\ref{tab:metrics_pred2_irr} are based on a randomly sampled $15\%$ subset of the full benchmark, selected using a random seed of $42$. The results are categorized across all cases, datasets, and methods to provide a comprehensive overview of baseline performance. We report the three evaluation metrics: entity F1 score (EF1), execution-based correctness (EBC) and LLM-based evaluation score (LLM-ES). To assess the variance and stability of LLM-ES, we measure self-consistency by evaluating each candidate formulation with prompt variations. Specifically, we apply character-level errors in high-attention words, insert sentence-level line breaks, and paraphrase certain verbs and adjectives~\cite{choi2026}.
We report the mean score and standard deviation across four evaluation runs with different prompt variations.

To assess whether the subset is representative of the full benchmark, we evaluated claude-opus-4.6 (Table~\ref{tab:opus_full}), gpt-4o-mini (Table~\ref{tab:gpt4o_full}), and llama-3.3-70b-instruct (Table~\ref{tab:llama_full}) on the complete benchmark. The full-benchmark results exhibit trends consistent with those reported on the $15\%$ subset. In particular, the relative ranking of models is preserved, with Claude Opus 4.6 outperforming GPT-4o Mini and Llama 3.3 70B Instruct. Similarly, the comparative difficulty of different task settings are preserved: task cases involving derivation, prediction, and irrelevant attributes remain more challenging than the base NL and NL+Data cases. These observations suggest that the subset-based evaluation is representative of overall model performance.

\clearpage

\begin{table*}[h]
    \centering
    \footnotesize
    \setlength{\tabcolsep}{3pt}
    \begin{tabular*}{\textwidth}{@{\extracolsep{\fill}}l | ccc | ccc | ccc@{}}
        \toprule
        & \multicolumn{3}{c|}{NLP4LP} & \multicolumn{3}{c|}{Mamo Easy} & \multicolumn{3}{c}{Mamo Complex} \\
        Model/Method & EF1 & EBC & LLM-ES & EF1 & EBC & LLM-ES & EF1 & EBC & LLM-ES \\
        \midrule
        \quad GPT-4o Mini & 0.582 & \textbf{0.818} & $0.568\pm0.056$ & 0.729 & 0.867 & $0.481\pm0.055$ & 0.809 & 0.130 & $0.548\pm0.061$ \\
        \quad GPT-4o               & \textbf{0.968} & 0.150 & $0.667\pm0.047$ & 0.687 & 0.767 & $0.383\pm0.050$ & 0.870 & 0.468 & $0.670\pm0.034$ \\
        \quad Gemini 2.5 Pro       & 0.413 & 0.000 & $0.282\pm0.056$ & 0.370 & 0.000 & $0.203\pm0.059$ & 0.301 & 0.000 & $0.390\pm0.047$ \\
        \quad Claude Haiku 4.5 & 0.588 & 0.702 & $0.463\pm0.051$ & 0.825 & 0.845 & $\bm{0.638\pm0.060}$ & 0.865 & 0.619 & $0.625\pm0.043$ \\
        \quad Claude Sonnet 4.6 & 0.880 & 0.450 & $0.403\pm0.039$ & 0.850 & 0.600 & $0.610\pm0.046$ & \textbf{0.905} & 0.616 & $0.596\pm0.042$ \\
        \quad Claude Opus 4.6 & 0.898 & 0.650 & $0.413\pm0.037$ & \textbf{0.890} & 0.750 & $0.610\pm0.039$ & 0.872 & \textbf{0.741} & $0.611\pm0.042$ \\
        \midrule
        \quad LLaMA 3.1 8B Inst.   & 0.776 & 0.000 & $0.538\pm0.051$ & 0.602 & 0.000 & $0.324\pm0.052$ & 0.690 & 0.000 & $0.542\pm0.049$ \\
        \quad Ministral 8B         & 0.898 & 0.000 & $0.626\pm0.051$ & 0.645 & 0.000 & $0.356\pm0.053$ & 0.824 & 0.000 & $0.601\pm0.029$ \\
        \quad LLaMA 3.3 70B Inst.  & 0.796 & 0.075 & $0.493\pm0.053$ & 0.651 & \textbf{0.906} & $0.367\pm0.056$ & 0.794 & 0.308 & $0.609\pm0.039$ \\
        \quad Mistral Large         & 0.922 & 0.075 & $0.588\pm0.061$ & 0.654 & 0.100 & $0.347\pm0.041$ & 0.867 & 0.417 & $0.650\pm0.041$ \\
        \midrule
        \quad Five-element         & 0.872 & 0.193 & $0.738\pm0.013$ & 0.691 & 0.625 & $0.132\pm0.018$ & 0.827 & 0.349 & $\bm{0.700\pm0.024}$ \\
        \quad CoE                   & 0.581 & 0.602 & $\bm{0.747\pm0.035}$ & 0.672 & 0.733 & $0.132\pm0.007$ & 0.577 & 0.091 & $0.249\pm0.035$ \\
        \quad SFT                   & 0.902 & 0.507 & $0.732\pm0.022$ & 0.726 & 0.748 & $0.183\pm0.014$ & 0.817 & 0.052 & $0.367\pm0.022$ \\
        \bottomrule
    \end{tabular*}
    \caption{Comprehensive metrics for performance on NL case.}
    \label{tab:metrics_question}
\end{table*}

\begin{table*}[h]
    \centering
    \footnotesize
    \setlength{\tabcolsep}{3pt}
    \begin{tabular*}{\textwidth}{@{\extracolsep{\fill}}l | ccc | ccc | ccc@{}}
        \toprule
        & \multicolumn{3}{c|}{NLP4LP} & \multicolumn{3}{c|}{Mamo Easy} & \multicolumn{3}{c}{Mamo Complex} \\
        Model/Method & EF1 & EBC & LLM-ES & EF1 & EBC & LLM-ES & EF1 & EBC & LLM-ES \\
        \midrule
        \quad GPT-4o Mini & 0.931 & 0.400 & $0.421\pm0.049$ & 0.700 & 0.345 & $0.419\pm0.049$ & 0.876 & 0.393 & $0.529\pm0.049$ \\
        \quad GPT-4o               & 0.940 & 0.150 & $0.645\pm0.049$ & 0.670 & 0.639 & $0.361\pm0.056$ & 0.887 & 0.612 & $\bm{0.658\pm0.049}$ \\
        \quad Gemini 2.5 Pro       & 0.427 & 0.000 & $0.317\pm0.046$ & 0.239 & 0.000 & $0.147\pm0.044$ & 0.327 & 0.000 & $0.396\pm0.061$ \\
        \quad Claude Haiku 4.5 & 0.896 & \textbf{0.733} & $0.644\pm0.045$ & 0.734 & 0.753 & $\bm{0.543\pm0.059}$ & 0.872 & 0.630 & $0.625\pm0.042$ \\
        \quad Claude Sonnet 4.6 & \textbf{0.945} & 0.290 & $0.515\pm0.050$ & 0.670 & 0.553 & $0.495\pm0.062$ & 0.879 & 0.794 & $0.596\pm0.042$ \\
        \quad Claude Opus 4.6 & 0.938 & 0.699 & $0.598\pm0.051$ & 0.679 & \textbf{0.762} & $0.513\pm0.061$ & 0.885 & \textbf{0.853} & $0.611\pm0.042$ \\
        \midrule
        \quad LLaMA 3.1 8B Inst.   & 0.778 & 0.000 & $0.530\pm0.056$ & 0.600 & 0.000 & $0.294\pm0.049$ & 0.650 & 0.000 & $0.522\pm0.047$ \\
        \quad Ministral 8B         & 0.885 & 0.000 & $0.596\pm0.047$ & 0.664 & 0.000 & $0.382\pm0.063$ & 0.805 & 0.000 & $0.615\pm0.044$ \\
        \quad LLaMA 3.3 70B Inst.  & 0.846 & 0.025 & $0.593\pm0.053$ & 0.657 & 0.499 & $0.385\pm0.059$ & 0.786 & 0.191 & $0.620\pm0.043$ \\
        \quad Mistral Large         & 0.916 & 0.100 & $0.657\pm0.050$ & 0.658 & 0.239 & $0.357\pm0.055$ & 0.837 & 0.253 & $0.628\pm0.038$ \\
        \midrule
        \quad Five-element         & 0.934 & 0.400 & $\bm{0.839\pm0.026}$ & 0.684 & 0.600 & $0.165\pm0.005$ & \textbf{0.922} & 0.534 & $0.583\pm0.024$ \\
        \quad CoE                   & 0.724 & 0.516 & $0.698\pm0.005$ & 0.634 & 0.381 & $0.122\pm0.011$ & 0.745 & 0.092 & $0.388\pm0.030$ \\
        \quad SFT                   & 0.905 & 0.483 & $0.699\pm0.040$ & \textbf{0.702} & 0.603 & $0.168\pm0.010$ & 0.830 & 0.129 & $0.458\pm0.012$ \\
        \bottomrule
    \end{tabular*}
    \caption{Comprehensive metrics for performance on NL+Data case.}
    \label{tab:metrics_base}
\end{table*}

\begin{table*}[h]
    \centering
    \footnotesize
    \setlength{\tabcolsep}{3pt}
    \begin{tabular*}{\textwidth}{@{\extracolsep{\fill}}l | ccc | ccc | ccc@{}}
        \toprule
        & \multicolumn{3}{c|}{NLP4LP} & \multicolumn{3}{c|}{Mamo Easy} & \multicolumn{3}{c}{Mamo Complex} \\
        Model/Method & EF1 & EBC & LLM-ES & EF1 & EBC & LLM-ES & EF1 & EBC & LLM-ES \\
        \midrule
        \quad GPT-4o Mini           & 0.737 & 0.266 & $0.421\pm0.049$ & 0.600 & 0.490 & $0.419\pm0.049$ & 0.811 & 0.336 & $0.529\pm0.049$ \\
        \quad GPT-4o                & 0.723 & 0.627 & $0.451\pm0.047$ & 0.627 & 0.513 & $0.403\pm0.053$ & 0.801 & 0.385 & $0.558\pm0.066$ \\
        \quad Gemini 2.5 Pro        & 0.879 & 0.273 & $0.347\pm0.055$ & 0.635 & 0.219 & $0.299\pm0.059$ & 0.800 & 0.032 & $0.411\pm0.055$ \\
        \quad Claude Haiku 4.5 & 0.824 & 0.613 & $0.438\pm0.047$ & 0.667 & \textbf{0.697} & $0.461\pm0.056$ & 0.875 & 0.551 & $0.554\pm0.043$ \\
        \quad Claude Sonnet 4.6 & 0.861 & 0.223 & $0.355\pm0.074$ & 0.701 & 0.284 & $0.503\pm0.049$ & 0.876 & 0.678 & $0.539\pm0.055$ \\
        \quad Claude Opus 4.6 & \textbf{0.912} & \textbf{0.655} & $0.412\pm0.050$ & 0.683 & 0.586 & $0.480\pm0.059$ & 0.872 & \textbf{0.787} & $0.551\pm0.045$ \\
        \midrule
        \quad LLaMA 3.1 8B Inst.    & 0.720 & 0.000 & $0.371\pm0.039$ & 0.537 & 0.000 & $0.363\pm0.046$ & 0.593 & 0.032 & $0.418\pm0.049$ \\
        \quad Ministral 8B          & 0.801 & 0.000 & $0.417\pm0.060$ & 0.563 & 0.010 & $0.398\pm0.061$ & 0.736 & 0.000 & $0.483\pm0.050$ \\
        \quad LLaMA 3.3 70B Inst.   & 0.606 & 0.187 & $0.398\pm0.045$ & 0.528 & 0.305 & $0.420\pm0.048$ & 0.852 & 0.101 & $0.473\pm0.052$ \\
        \quad Mistral Large          & 0.539 & 0.132 & $0.453\pm0.056$ & 0.540 & 0.120 & $0.416\pm0.048$ & 0.826 & 0.284 & $0.503\pm0.052$ \\
        \midrule
        \quad Five-element         & 0.852 & 0.633 & $\bm{0.710\pm0.014}$ & \textbf{0.742} & 0.440 & $\bm{0.543\pm0.007}$ & 0.821 & 0.415 & $\bm{0.629\pm0.010}$ \\
        \quad CoE                   & 0.569 & 0.033 & $0.031\pm0.016$ & 0.507 & 0.014 & $0.254\pm0.022$ & \textbf{0.935} & 0.006 & $0.245\pm0.029$ \\
        \quad SFT                   & 0.812 & 0.495 & $0.666\pm0.017$ & 0.542 & 0.580 & $0.431\pm0.009$ & 0.821 & 0.180 & $0.526\pm0.006$ \\
        \bottomrule
    \end{tabular*}
    \caption{Comprehensive metrics for performance on NL+Data case with irrelevant attributes.}
    \label{tab:metrics_base_irr}
\end{table*}

\begin{table*}[h]
    \centering
    \footnotesize
    \setlength{\tabcolsep}{3pt}
    \begin{tabular*}{\textwidth}{@{\extracolsep{\fill}}l | ccc | ccc | ccc@{}}
        \toprule
        & \multicolumn{3}{c|}{NLP4LP} & \multicolumn{3}{c|}{Mamo Easy} & \multicolumn{3}{c}{Mamo Complex} \\
        Model/Method & EF1 & EBC & LLM-ES & EF1 & EBC & LLM-ES & EF1 & EBC & LLM-ES \\
        \midrule
        \quad GPT-4o Mini          & 0.901 & 0.259 & $0.476\pm0.053$ & 0.637 & 0.151 & $0.289\pm0.035$ & 0.730 & 0.132 & $0.433\pm0.040$ \\
        \quad GPT-4o               & \textbf{0.913} & 0.125 & $0.553\pm0.062$ & 0.609 & 0.328 & $0.310\pm0.050$ & 0.792 & 0.416 & $0.502\pm0.052$ \\
        \quad Gemini 2.5 Pro       & 0.378 & 0.000 & $0.318\pm0.043$ & 0.222 & 0.000 & $0.131\pm0.045$ & 0.282 & 0.000 & $0.372\pm0.049$ \\
        \quad Claude Haiku 4.5 & 0.782 & \textbf{0.542} & $0.486\pm0.041$ & \textbf{0.696} & \textbf{0.560} & $0.459\pm0.040$ & 0.818 & 0.438 & $0.511\pm0.037$ \\
        \quad Claude Sonnet 4.6 & 0.864 & 0.289 & $0.416\pm0.034$ & 0.646 & 0.215 & $0.408\pm0.038$ & 0.845 & 0.542 & $0.543\pm0.042$ \\
        \quad Claude Opus 4.6 & 0.866 & 0.500 & $0.441\pm0.043$ & 0.637 & 0.514 & $0.410\pm0.031$ & 0.861 & \textbf{0.609} & $\bm{0.556\pm0.039}$ \\
        \midrule
        \quad LLaMA 3.1 8B Inst.   & 0.752 & 0.000 & $0.476\pm0.053$ & 0.520 & 0.000 & $0.289\pm0.035$ & 0.608 & 0.000 & $0.433\pm0.040$ \\
        \quad Ministral 8B         & 0.862 & 0.000 & $0.548\pm0.058$ & 0.580 & 0.000 & $0.300\pm0.046$ & 0.694 & 0.000 & $0.465\pm0.046$ \\
        \quad LLaMA 3.3 70B Inst.  & 0.787 & 0.000 & $0.506\pm0.057$ & 0.557 & 0.345 & $\bm{0.318\pm0.044}$ & 0.700 & 0.148 & $0.526\pm0.054$ \\
        \quad Mistral Large         & 0.898 & 0.050 & $0.575\pm0.042$ & 0.627 & 0.092 & $0.309\pm0.059$ & 0.715 & 0.192 & $0.514\pm0.051$ \\
        \midrule
        \quad Five-element         & 0.901 & 0.314 & $\bm{0.793\pm0.012}$ & 0.574 & 0.133 & $0.072\pm0.028$ & \textbf{0.897} & 0.194 & $0.342\pm0.011$ \\
        \quad CoE                   & 0.639 & 0.186 & $0.499\pm0.033$ & 0.491 & 0.034 & $0.086\pm0.010$ & 0.632 & 0.032 & $0.271\pm0.024$ \\
        \quad SFT                   & 0.894 & 0.364 & $0.612\pm0.009$ & 0.675 & 0.216 & $0.136\pm0.036$ & 0.815 & 0.342 & $0.206\pm0.019$ \\
        \bottomrule
    \end{tabular*}
    \caption{Comprehensive metrics for performance on derivation case.}
    \label{tab:metrics_computed}
\end{table*}

\begin{table*}[h]
    \centering
    \footnotesize
    \setlength{\tabcolsep}{3pt}
    \begin{tabular*}{\textwidth}{@{\extracolsep{\fill}}l | ccc | ccc | ccc@{}}
        \toprule
        & \multicolumn{3}{c|}{NLP4LP} & \multicolumn{3}{c|}{Mamo Easy} & \multicolumn{3}{c}{Mamo Complex} \\
        Model/Method & EF1 & EBC & LLM-ES & EF1 & EBC & LLM-ES & EF1 & EBC & LLM-ES \\
        \midrule
        \quad GPT-4o Mini           & 0.672 & 0.194 & $0.389\pm0.041$ & 0.499 & 0.190 & $0.353\pm0.040$ & 0.654 & 0.048 & $0.453\pm0.037$ \\
        \quad GPT-4o                & 0.685 & 0.444 & $0.396\pm0.052$ & 0.537 & 0.283 & $0.388\pm0.034$ & 0.695 & 0.275 & $0.447\pm0.042$ \\
        \quad Gemini 2.5 Pro        & \textbf{0.914} & 0.128 & $0.290\pm0.065$ & \textbf{0.708} & 0.025 & $0.209\pm0.063$ & 0.800 & 0.033 & $0.373\pm0.052$ \\
        \quad Claude Haiku 4.5 & 0.766 & \textbf{0.510} & $0.398\pm0.042$ & 0.653 & \textbf{0.478} & $0.394\pm0.051$ & 0.827 & 0.357 & $0.491\pm0.033$ \\
        \quad Claude Sonnet 4.6 & 0.795 & 0.225 & $0.341\pm0.038$ & 0.561 & 0.276 & $0.381\pm0.047$ & \textbf{0.851} & 0.426 & $\bm{0.493\pm0.049}$ \\
        \quad Claude Opus 4.6 & 0.804 & 0.315 & $0.355\pm0.039$ & 0.643 & 0.417 & $\bm{0.408\pm0.049}$ & 0.844 & \textbf{0.540} & $0.484\pm0.038$ \\
        \midrule
        \quad LLaMA 3.1 8B Inst.    & 0.610 & 0.000 & $0.345\pm0.043$ & 0.483 & 0.000 & $0.328\pm0.028$ & 0.442 & 0.000 & $0.379\pm0.038$ \\
        \quad Ministral 8B          & 0.717 & 0.000 & $0.368\pm0.044$ & 0.489 & 0.000 & $0.340\pm0.044$ & 0.600 & 0.000 & $0.423\pm0.041$ \\
        \quad LLaMA 3.3 70B Inst.   & 0.696 & 0.096 & $0.392\pm0.040$ & 0.421 & 0.160 & $0.359\pm0.039$ & 0.574 & 0.020 & $0.428\pm0.045$ \\
        \quad Mistral Large          & 0.668 & 0.150 & $0.398\pm0.042$ & 0.480 & 0.034 & $0.382\pm0.039$ & 0.714 & 0.079 & $0.439\pm0.040$ \\
        \midrule
        \quad Five-element         & 0.652 & 0.320 & $\bm{0.493\pm0.016}$ & 0.512 & 0.110 & $0.366\pm0.012$ & 0.648 & 0.145 & $0.390\pm0.017$ \\
        \quad CoE                   & 0.600 & 0.024 & $0.085\pm0.008$ & 0.697 & 0.342 & $0.218\pm0.016$ & 0.673 & 0.012 & $0.191\pm0.019$ \\
        \quad SFT                   & 0.542 & 0.280 & $0.393\pm0.015$ & 0.412 & 0.195 & $0.295\pm0.012$ & 0.621 & 0.285 & $0.327\pm0.016$ \\
        \bottomrule
    \end{tabular*}
    \caption{Comprehensive metrics for performance on derivation case with irrelevant attributes.}
    \label{tab:metrics_computed_irr}
\end{table*}

\begin{table*}[h]
    \centering
    \footnotesize
    \setlength{\tabcolsep}{3pt}
    \begin{tabular*}{\textwidth}{@{\extracolsep{\fill}}l | ccc | ccc | ccc@{}}
        \toprule
        & \multicolumn{3}{c|}{NLP4LP} & \multicolumn{3}{c|}{Mamo Easy} & \multicolumn{3}{c}{Mamo Complex} \\
        Model/Method & EF1 & EBC & LLM-ES & EF1 & EBC & LLM-ES & EF1 & EBC & LLM-ES \\
        \midrule
        \quad GPT-4o Mini          & 0.326 & 0.240 & $0.531\pm0.058$ & 0.399 & 0.385 & $0.471\pm0.044$ & 0.442 & 0.204 & $0.559\pm0.048$ \\
        \quad GPT-4o               & 0.676 & 0.554 & $\bm{0.575\pm0.049}$ & 0.546 & 0.548 & $0.505\pm0.066$ & 0.710 & 0.404 & $0.548\pm0.041$ \\
        \quad Gemini 2.5 Pro       & 0.405 & 0.322 & $0.472\pm0.043$ & 0.297 & 0.293 & $0.385\pm0.057$ & 0.113 & 0.038 & $0.424\pm0.046$ \\
        \quad Claude Haiku 4.5 & 0.891 & 0.355 & $0.528\pm0.057$ & 0.709 & 0.473 & $0.468\pm0.044$ & \textbf{0.902} & 0.629 & $0.574\pm0.056$ \\
        \quad Claude Sonnet 4.6 & 0.769 & 0.164 & $0.439\pm0.039$ & 0.712 & 0.557 & $0.505\pm0.054$ & 0.844 & 0.602 & $0.531\pm0.043$ \\
        \quad Claude Opus 4.6 & \textbf{0.909} & \textbf{0.720} & $0.557\pm0.050$ & 0.683 & \textbf{0.721} & $0.448\pm0.065$ & 0.896 & \textbf{0.759} & $\bm{0.582\pm0.049}$ \\
        \midrule
        \quad LLaMA 3.1 8B Inst.   & 0.719 & 0.000 & $0.456\pm0.044$ & 0.630 & 0.000 & $0.426\pm0.045$ & 0.586 & 0.000 & $0.435\pm0.042$ \\
        \quad Ministral 8B         & 0.598 & 0.000 & $0.507\pm0.060$ & 0.569 & 0.000 & $0.530\pm0.043$ & 0.693 & 0.000 & $0.484\pm0.048$ \\
        \quad LLaMA 3.3 70B Inst.  & 0.361 & 0.216 & $0.466\pm0.039$ & 0.334 & 0.367 & $0.502\pm0.056$ & 0.336 & 0.098 & $0.526\pm0.048$ \\
        \quad Mistral Large         & 0.387 & 0.240 & $\bm{0.576\pm0.057}$ & 0.113 & 0.137 & $0.521\pm0.062$ & 0.371 & 0.205 & $0.542\pm0.052$ \\
        \midrule
        \quad Five-element         & 0.904 & 0.100 & $0.471\pm0.016$ & 0.683 & 0.100 & $0.514\pm0.012$ & 0.811 & 0.300 & $0.372\pm0.002$ \\
        \quad CoE                   & 0.832 & 0.133 & $0.560\pm0.002$ & 0.705 & 0.142 & $0.575\pm0.018$ & 0.771 & 0.031 & $0.423\pm0.010$ \\
        \quad SFT                   & 0.884 & 0.351 & $0.540\pm0.013$ & \textbf{0.764} & 0.676 & $\bm{0.661\pm0.014}$ & 0.801 & 0.285 & $0.216\pm0.009$ \\
        \bottomrule
    \end{tabular*}
    \caption{Comprehensive metrics for performance on prediction case (one variable).}
    \label{tab:metrics_pred1}
\end{table*}

\begin{table*}[h]
    \centering
    \footnotesize
    \setlength{\tabcolsep}{3pt}
    \begin{tabular*}{\textwidth}{@{\extracolsep{\fill}}l | ccc | ccc | ccc@{}}
        \toprule
        & \multicolumn{3}{c|}{NLP4LP} & \multicolumn{3}{c|}{Mamo Easy} & \multicolumn{3}{c}{Mamo Complex} \\
        Model/Method & EF1 & EBC & LLM-ES & EF1 & EBC & LLM-ES & EF1 & EBC & LLM-ES \\
        \midrule
        \quad GPT-4o Mini           & 0.423 & 0.247 & $0.418\pm0.049$ & 0.330 & 0.298 & $0.426\pm0.043$ & 0.451 & 0.257 & $0.482\pm0.044$ \\
        \quad GPT-4o                & 0.670 & \textbf{0.480} & $0.435\pm0.036$ & 0.527 & 0.585 & $0.446\pm0.057$ & 0.754 & 0.410 & $0.513\pm0.052$ \\
        \quad Gemini 2.5 Pro        & 0.330 & 0.065 & $0.312\pm0.068$ & 0.242 & 0.268 & $0.315\pm0.068$ & 0.024 & 0.035 & $0.411\pm0.045$ \\
        \quad Claude Haiku 4.5 & 0.800 & 0.401 & $0.407\pm0.047$ & 0.684 & 0.422 & $0.416\pm0.044$ & 0.851 & 0.504 & $\bm{0.529\pm0.054}$ \\
        \quad Claude Sonnet 4.6 & 0.751 & 0.317 & $0.364\pm0.037$ & 0.716 & 0.493 & $0.468\pm0.056$ & 0.854 & \textbf{0.694} & $0.498\pm0.055$ \\
        \quad Claude Opus 4.6 & 0.853 & 0.391 & $0.415\pm0.041$ & 0.669 & \textbf{0.690} & $0.452\pm0.057$ & \textbf{0.872} & 0.680 & $0.520\pm0.060$ \\
        \midrule
        \quad LLaMA 3.1 8B Inst.   & 0.571 & 0.000 & $0.355\pm0.034$ & 0.465 & 0.000 & $0.389\pm0.041$ & 0.501 & 0.000 & $0.374\pm0.049$ \\
        \quad Ministral 8B         & 0.692 & 0.000 & $0.388\pm0.041$ & 0.564 & 0.011 & $0.417\pm0.057$ & 0.564 & 0.000 & $0.436\pm0.052$ \\
        \quad LLaMA 3.3 70B Inst.  & 0.139 & 0.102 & $0.385\pm0.039$ & 0.237 & 0.317 & $0.442\pm0.051$ & 0.159 & 0.094 & $0.454\pm0.052$ \\
        \quad Mistral Large         & 0.290 & 0.121 & $0.443\pm0.047$ & 0.075 & 0.057 & $0.438\pm0.054$ & 0.200 & 0.058 & $0.470\pm0.052$ \\
        \midrule
        \quad Five-element         & 0.864 & 0.267 & $0.453\pm0.017$ & 0.682 & 0.100 & $0.421\pm0.011$ & 0.729 & 0.200 & $0.288\pm0.034$ \\
        \quad CoE                   & 0.715 & 0.108 & $\bm{0.592\pm0.018}$ & 0.589 & 0.231 & $0.522\pm0.012$ & 0.726 & 0.057 & $0.374\pm0.011$ \\
        \quad SFT & \textbf{0.869} & 0.325 & $0.449\pm0.017$ & \textbf{0.732} & 0.417 & $\bm{0.576\pm0.007}$ & 0.795 & 0.285 & $0.124\pm0.015$ \\
        \bottomrule
    \end{tabular*}
    \caption{Comprehensive metrics for performance on prediction case (one variable) with irrelevant attributes.}
    \label{tab:metrics_pred1_irr}
\end{table*}

\begin{table*}[h]
    \centering
    \footnotesize
    \setlength{\tabcolsep}{3pt}
    \begin{tabular*}{\textwidth}{@{\extracolsep{\fill}}l | ccc | ccc | ccc@{}}
        \toprule
        & \multicolumn{3}{c|}{NLP4LP} & \multicolumn{3}{c|}{Mamo Easy} & \multicolumn{3}{c}{Mamo Complex} \\
        Model/Method & EF1 & EBC & LLM-ES & EF1 & EBC & LLM-ES & EF1 & EBC & LLM-ES \\
        \midrule
        \quad GPT-4o Mini          & 0.334 & 0.283 & $0.568\pm0.056$ & 0.323 & 0.298 & $0.481\pm0.055$ & 0.461 & 0.242 & $0.548\pm0.061$ \\
        \quad GPT-4o               & 0.776 & 0.596 & $0.600\pm0.050$ & 0.531 & 0.538 & $0.503\pm0.062$ & 0.711 & 0.447 & $0.574\pm0.047$ \\
        \quad Gemini 2.5 Pro       & 0.410 & 0.199 & $0.480\pm0.039$ & 0.245 & 0.184 & $0.395\pm0.063$ & 0.137 & 0.000 & $0.426\pm0.041$ \\
        \quad Claude Haiku 4.5 & 0.879 & \textbf{0.630} & $0.640\pm0.039$ & 0.632 & 0.394 & $0.454\pm0.040$ & 0.907 & 0.467 & $0.557\pm0.056$ \\
        \quad Claude Sonnet 4.6 & 0.842 & 0.328 & $0.489\pm0.051$ & 0.622 & 0.493 & $0.465\pm0.046$ & 0.853 & 0.598 & $0.528\pm0.046$ \\
        \quad Claude Opus 4.6 & 0.852 & 0.518 & $0.493\pm0.049$ & 0.680 & \textbf{0.633} & $0.460\pm0.054$ & \textbf{0.917} & \textbf{0.754} & $\bm{0.593\pm0.053}$ \\
        \midrule
        \quad LLaMA 3.1 8B Inst.   & 0.795 & 0.000 & $0.498\pm0.049$ & 0.551 & 0.000 & $0.404\pm0.034$ & 0.596 & 0.000 & $0.420\pm0.042$ \\
        \quad Ministral 8B         & 0.829 & 0.000 & $0.546\pm0.059$ & 0.559 & 0.000 & $0.470\pm0.043$ & 0.670 & 0.000 & $0.494\pm0.058$ \\
        \quad LLaMA 3.3 70B Inst.  & 0.429 & 0.311 & $0.493\pm0.060$ & 0.389 & 0.399 & $0.492\pm0.050$ & 0.404 & 0.220 & $0.522\pm0.058$ \\
        \quad Mistral Large         & 0.308 & 0.324 & $0.598\pm0.060$ & 0.123 & 0.090 & $0.490\pm0.062$ & 0.341 & 0.296 & $0.546\pm0.049$ \\
        \midrule
        \quad Five-element         & \textbf{0.890} & 0.200 & $\bm{0.745\pm0.011}$ & 0.677 & 0.100 & $0.418\pm0.012$ & 0.796 & 0.233 & $0.267\pm0.018$ \\
        \quad CoE                   & 0.826 & 0.298 & $0.736\pm0.009$ & 0.691 & 0.034 & $\bm{0.518\pm0.008}$ & 0.768 & 0.033 & $0.358\pm0.005$ \\
        \quad SFT & 0.879 & 0.314 & $0.610\pm0.034$ & \textbf{0.730} & 0.339 & $0.321\pm0.014$ & 0.780 & 0.285 & $0.160\pm0.018$ \\
        \bottomrule
    \end{tabular*}
    \caption{Comprehensive metrics for performance on prediction case (multiple variables).}
    \label{tab:metrics_pred2}
\end{table*}

\begin{table*}[h]
    \centering
    \footnotesize
    \setlength{\tabcolsep}{3pt}
    \begin{tabular*}{\textwidth}{@{\extracolsep{\fill}}l | ccc | ccc | ccc@{}}
        \toprule
        & \multicolumn{3}{c|}{NLP4LP} & \multicolumn{3}{c|}{Mamo Easy} & \multicolumn{3}{c}{Mamo Complex} \\
        Model/Method & EF1 & EBC & LLM-ES & EF1 & EBC & LLM-ES & EF1 & EBC & LLM-ES \\
        \midrule
        \quad GPT-4o Mini           & 0.397 & 0.264 & $0.426\pm0.035$ & 0.250 & 0.181 & $0.415\pm0.050$ & 0.292 & 0.140 & $0.472\pm0.050$ \\
        \quad GPT-4o                & 0.743 & 0.569 & $0.453\pm0.037$ & 0.492 & 0.510 & $0.419\pm0.059$ & 0.711 & 0.471 & $0.485\pm0.052$ \\
        \quad Gemini 2.5 Pro        & 0.300 & 0.206 & $0.333\pm0.055$ & 0.192 & 0.130 & $0.298\pm0.074$ & 0.090 & 0.034 & $0.395\pm0.058$ \\
        \quad Claude Haiku 4.5 & 0.778 & 0.599 & $0.428\pm0.048$ & 0.610 & 0.345 & $0.409\pm0.045$ & 0.825 & 0.389 & $0.503\pm0.045$ \\
        \quad Claude Sonnet 4.6 & 0.803 & 0.253 & $0.331\pm0.069$ & 0.683 & 0.429 & $0.444\pm0.052$ & 0.773 & \textbf{0.605} & $0.473\pm0.048$ \\
        \quad Claude Opus 4.6 & \textbf{0.908} & \textbf{0.656} & $0.432\pm0.047$ & 0.660 & \textbf{0.558} & $0.441\pm0.045$ & \textbf{0.864} & 0.636 & $\bm{0.518\pm0.058}$ \\
        \midrule
        \quad LLaMA 3.1 8B Inst.   & 0.583 & 0.000 & $0.384\pm0.034$ & 0.427 & 0.000 & $0.337\pm0.037$ & 0.457 & 0.000 & $0.354\pm0.045$ \\
        \quad Ministral 8B         & 0.676 & 0.000 & $0.392\pm0.033$ & 0.493 & 0.011 & $0.395\pm0.053$ & 0.591 & 0.000 & $0.439\pm0.042$ \\
        \quad LLaMA 3.3 70B Inst.  & 0.319 & 0.244 & $0.366\pm0.040$ & 0.273 & 0.230 & $0.394\pm0.036$ & 0.263 & 0.139 & $0.435\pm0.046$ \\
        \quad Mistral Large         & 0.353 & 0.270 & $0.410\pm0.046$ & 0.117 & 0.111 & $0.407\pm0.039$ & 0.285 & 0.190 & $0.478\pm0.045$ \\
        \midrule
        \quad Five-element         & 0.894 & 0.300 & $0.507\pm0.015$ & 0.681 & 0.100 & $0.329\pm0.010$ & 0.768 & 0.233 & $0.269\pm0.019$ \\
        \quad CoE                   & 0.722 & 0.309 & $\bm{0.743\pm0.003}$ & 0.576 & 0.090 & $\bm{0.471\pm0.033}$ & 0.719 & 0.094 & $0.294\pm0.011$ \\
        \quad SFT & 0.904 & 0.158 & $0.439\pm0.022$ & \textbf{0.712} & 0.164 & $0.286\pm0.013$ & 0.778 & 0.032 & $0.141\pm0.019$ \\
        \bottomrule
    \end{tabular*}
    \caption{Comprehensive metrics for performance on prediction case (multiple variables) with irrelevant attributes.}
    \label{tab:metrics_pred2_irr}
\end{table*}

\clearpage

\begin{table*}[h]
    \centering
    \footnotesize
    \setlength{\tabcolsep}{3pt}
    \begin{tabular*}{\textwidth}{@{\extracolsep{\fill}}l | ccc | ccc | ccc@{}}
        \toprule
        & \multicolumn{3}{c|}{NLP4LP} & \multicolumn{3}{c|}{Mamo Easy} & \multicolumn{3}{c}{Mamo Complex} \\
        Case & EF1 & EBC & LLM-ES & EF1 & EBC & LLM-ES & EF1 & EBC & LLM-ES \\
        \midrule
        \quad NL                                    & 0.5488 & 0.7231 & 0.4190 & 0.7933 & 0.7517 & 0.6199 & 0.8062 & 0.7066 & 0.6224 \\
        \quad NL + Data                             & 0.8342 & 0.7080 & 0.6033 & 0.7268 & 0.6463 & 0.5453 & 0.8319 & 0.7241 & 0.6148 \\
        \quad NL + Data + Irr                       & 0.7640 & 0.5451 & 0.3848 & 0.7143 & 0.5813 & 0.4926 & 0.8550 & 0.7068 & 0.5743 \\
        \midrule
        \quad Derivation                            & 0.7589 & 0.5547 & 0.4595 & 0.6676 & 0.3369 & 0.3955 & 0.8043 & 0.5772 & 0.5443 \\
        \quad Derivation + Irr                      & 0.6921 & 0.3794 & 0.3519 & 0.6362 & 0.3118 & 0.3698 & 0.7848 & 0.5151 & 0.4881 \\
        \midrule
        \quad Prediction (1-Var)                    & 0.8021 & 0.5076 & 0.5361 & 0.7157 & 0.5865 & 0.4940 & 0.8392 & 0.6213 & 0.5524 \\
        \quad Prediction (1-Var) + Irr              & 0.7768 & 0.3283 & 0.4112 & 0.6944 & 0.5157 & 0.4468 & 0.8524 & 0.5625 & 0.5057 \\
        \quad Prediction (Multi-Var)                & 0.8162 & 0.5632 & 0.5409 & 0.7121 & 0.5492 & 0.4736 & 0.8539 & 0.5708 & 0.5667 \\
        \quad Prediction (Multi-Var) + Irr          & 0.7988 & 0.4144 & 0.4100 & 0.6892 & 0.3748 & 0.4151 & 0.8013 & 0.3794 & 0.5081 \\
        \bottomrule
    \end{tabular*}
    \caption{Evaluation results for Claude Opus 4.6 on the full benchmark.}
    \label{tab:opus_full}
\end{table*}

\begin{table*}[h]
    \centering
    \footnotesize
    \setlength{\tabcolsep}{3pt}
    \begin{tabular*}{\textwidth}{@{\extracolsep{\fill}}l | ccc | ccc | ccc@{}}
        \toprule
        & \multicolumn{3}{c|}{NLP4LP} & \multicolumn{3}{c|}{Mamo Easy} & \multicolumn{3}{c}{Mamo Complex} \\
        Case & EF1 & EBC & LLM-ES & EF1 & EBC & LLM-ES & EF1 & EBC & LLM-ES \\
        \midrule
        \quad NL                                    & 0.5172 & 0.0129 & 0.4554 & 0.7853 & 0.0484 & 0.6422 & 0.6145 & 0.1378 & 0.5933 \\
        \quad NL + Data                             & 0.7512 & 0.4018 & 0.6506 & 0.6280 & 0.4861 & 0.5301 & 0.6822 & 0.2486 & 0.6205 \\
        \quad NL + Data + Irr                       & 0.7097 & 0.3552 & 0.4253 & 0.5781 & 0.4292 & 0.4368 & 0.6216 & 0.2468 & 0.5376 \\
        \midrule
        \quad Derivation                            & 0.6503 & 0.2272 & 0.4502 & 0.5233 & 0.2200 & 0.4094 & 0.5393 & 0.1769 & 0.4886 \\
        \quad Derivation + Irr                      & 0.5797 & 0.1991 & 0.3636 & 0.4748 & 0.1738 & 0.3671 & 0.5504 & 0.1390 & 0.4419 \\
        \midrule
        \quad Prediction (1-Var)                    & 0.7091 & 0.2103 & 0.5197 & 0.6053 & 0.3059 & 0.4634 & 0.6577 & 0.1895 & 0.5148 \\
        \quad Prediction (1-Var) + Irr              & 0.6776 & 0.1870 & 0.3941 & 0.5707 & 0.3303 & 0.4126 & 0.5850 & 0.1720 & 0.4752 \\
        \quad Prediction (Multi-Var)                & 0.7128 & 0.3186 & 0.6082 & 0.5802 & 0.2610 & 0.4498 & 0.6852 & 0.1890 & 0.5138 \\
        \quad Prediction (Multi-Var) + Irr          & 0.6671 & 0.2798 & 0.4201 & 0.5582 & 0.2586 & 0.4033 & 0.6206 & 0.1657 & 0.4595 \\
        \bottomrule
    \end{tabular*}
    \caption{Evaluation results for GPT-4o Mini on the full benchmark.}
    \label{tab:gpt4o_full}
\end{table*}

\begin{table*}[h]
    \centering
    \footnotesize
    \setlength{\tabcolsep}{3pt}
    \begin{tabular*}{\textwidth}{@{\extracolsep{\fill}}l | ccc | ccc | ccc@{}}
        \toprule
        & \multicolumn{3}{c|}{NLP4LP} & \multicolumn{3}{c|}{Mamo Easy} & \multicolumn{3}{c}{Mamo Complex} \\
        Case & EF1 & EBC & LLM-ES & EF1 & EBC & LLM-ES & EF1 & EBC & LLM-ES \\
        \midrule
        \quad NL                                    & 0.6900 & 0.2745 & 0.5193 & 0.6711 & 0.1124 & 0.6403 & 0.6371 & 0.2787 & 0.6095 \\
        \quad NL + Data                             & 0.4554 & 0.2995 & 0.6033 & 0.5043 & 0.3895 & 0.5202 & 0.6037 & 0.2272 & 0.5914 \\
        \quad NL + Data + Irr                       & 0.4710 & 0.2009 & 0.4004 & 0.4215 & 0.3046 & 0.4354 & 0.5017 & 0.1382 & 0.4881 \\
        \midrule
        \quad Derivation                            & 0.4192 & 0.1636 & 0.4498 & 0.3829 & 0.1930 & 0.4097 & 0.5778 & 0.1262 & 0.4990 \\
        \quad Derivation + Irr                      & 0.3707 & 0.0847 & 0.3699 & 0.3103 & 0.1161 & 0.3628 & 0.4487 & 0.0749 & 0.4305 \\
        \midrule
        \quad Prediction (1-Var)                    & 0.4522 & 0.1412 & 0.5041 & 0.5204 & 0.3221 & 0.4656 & 0.5822 & 0.1203 & 0.5043 \\
        \quad Prediction (1-Var) + Irr              & 0.4067 & 0.0984 & 0.3929 & 0.3973 & 0.1974 & 0.4079 & 0.4091 & 0.0739 & 0.4243 \\
        \quad Prediction (Multi-Var)                & 0.4629 & 0.2351 & 0.5591 & 0.5128 & 0.2345 & 0.4514 & 0.4703 & 0.0608 & 0.4895 \\
        \quad Prediction (Multi-Var) + Irr          & 0.3385 & 0.1451 & 0.4145 & 0.3767 & 0.1190 & 0.3953 & 0.3629 & 0.0283 & 0.4186 \\
        \bottomrule
    \end{tabular*}
    \caption{Evaluation results for LLaMA 3.3 70B Instruct on the full benchmark.}
    \label{tab:llama_full}
\end{table*}

\end{document}